\documentclass[runningheads]{llncs}

\usepackage{eccv}

\usepackage{eccvabbrv}
\usepackage{graphicx}
\usepackage{booktabs}
\usepackage{multirow}
\usepackage{makecell}
\usepackage{algorithm}
\usepackage{algpseudocode}
\usepackage{wrapfig}
\usepackage[table]{xcolor}

\usepackage[accsupp]{axessibility}  

\usepackage{hyperref}

\begin{document}

\title{G2P: Gaussian-to-Point Attribute Alignment for Boundary-Aware 3D Segmentation} 

\titlerunning{G2P: Gaussian-to-Point Attribute Alignment}

\author{%
Hojun Song\inst{1}$^{*}$
\and
Chae-yeong Song\inst{1,2}$^{*,\ddagger}$
\and
Jeong-hun Hong\inst{1}
\and
Chaewon Moon\inst{1}
\and \\
Soo Ye Kim\inst{3}
\and
Yiyi Liao\inst{4}
\and
Jaehyup Lee\inst{1}
\and
Sang-hyo Park\inst{1}$^{\dagger}$
}

\authorrunning{H.~Song et al.}

\institute{%
Kyungpook National University
\and
Korea Electronics Technology Institute
\and
Adobe Research
\and
Zhejiang University \\
\url{https://hojunking.github.io/webpages/G2P/}
}

\maketitle
\renewcommand{\thefootnote}{}
\footnotetext{$^{*}$~Equal contribution.\quad$^{\dagger}$~Corresponding author.}
\footnotetext{$^{\ddagger}$~Work done while at Kyungpook National University.}
\renewcommand{\thefootnote}{\arabic{footnote}}

\begin{abstract}
  Point cloud segmentation is critical for 3D scene understanding. However, sparse and irregular point distributions provide limited appearance evidence, making geometry-only features insufficient to distinguish objects with similar shapes but distinct appearances (\eg, color, texture, and material). We propose Gaussian-to-Point (G2P), which transfers Gaussian attributes from 3D Gaussian Splatting to point clouds for more discriminative and appearance-consistent segmentation. Our G2P addresses the misalignment between optimized Gaussians and original point geometry by establishing point-wise correspondences. By distilling opacity-derived visibility cues, we mitigate the geometric ambiguity that limits existing models. Additionally, Gaussian scale attributes enable precise boundary localization in complex 3D scenes. Extensive experiments demonstrate that our approach achieves competitive performance on standard benchmarks and shows notable improvements on geometrically challenging classes, without pretrained 2D features or language supervision in our segmentation pipeline.
\end{abstract}

\section{Introduction}
\label{sec:intro}

\definecolor{hiPink}{RGB}{255,80,160}   
\definecolor{hiYellow}{RGB}{255,210,0}  

\begin{figure}[t]
    \centering
    \includegraphics[width=0.69\linewidth]{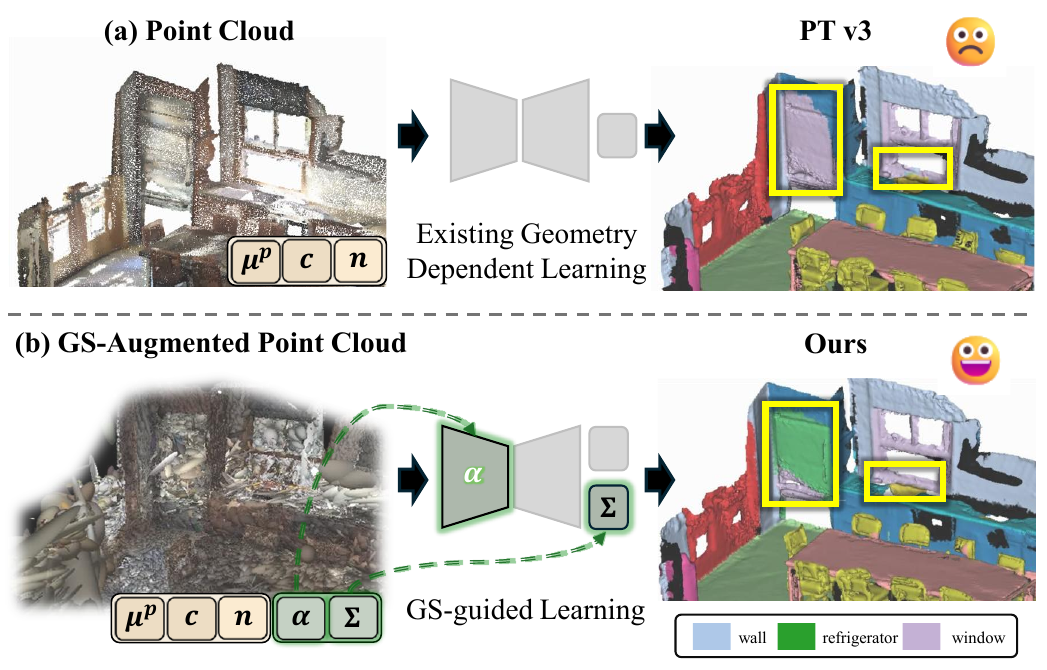}
    \caption{\textbf{G2P augments points with Gaussian attributes while preserving geometry.}
(a) \textit{Point Cloud.} Traditional point cloud representations lack sufficient appearance information to distinguish objects with similar geometry. As shown in the prediction, the existing model fails to segment coplanar windows cleanly and misclassifies the metallic refrigerator (\textcolor{hiYellow}{\textbf{yellow}} boxes), revealing geometric bias from insufficient appearance cues. (b) \textit{GS-Augmented Point Cloud (ours).} Through our novel Gaussian-to-Point feature augmentation, G2P enriches points with Gaussian attributes $(\Sigma, \alpha)$, which provide Gaussian-derived visibility and scale cues. This enables successful segmentation of both challenging cases, mitigating the geometric bias problem.
}
    \label{fig:1}
\end{figure}

Point cloud segmentation is a fundamental task that enables comprehensive 3D scene understanding across diverse real-world applications. Despite strong progress~\cite{Minkwski, oct, PointTFv3, oneformer3d}, the intrinsic sparsity and irregular sampling of point clouds force models to over-rely on coarse geometry, leading to geometric bias that confuses objects with similar shapes but distinct appearances~\cite{odin}. For instance, background-adjacent objects (\eg, doors, windows, and refrigerators) that are coplanar with walls or floors often become indistinguishable from their surroundings in sparse point clouds (\cref{fig:1}(a)). This geometric bias highlights the need to align appearance cues for accurate discrimination.

Recent advances have improved geometric understanding of point clouds, but existing approaches still face two fundamental limitations. \textbf{1) Boundary ambiguity.} Boundary-aware approaches~\cite{jsenet, tang2022contrastive, bfanet} explicitly model edges, refining object boundaries. However, they remain limited to geometric reasoning and do not take advantage of the appearance information essential to address geometric bias~\cite{dai2017bundlefusion, 3dpc-survey}. \textbf{2) Cross-modal misalignment.} 2D-3D fusion approaches~\cite{multiview,jaritz2019multiview, chiang2019unified, robert2022learning, vdg-uni3dseg} inject rich image features but suffer from structural mismatches~\cite{3dpc-survey}. The discrete nature of point clouds makes wire-like structures hard to distinguish from flat surfaces~\cite{mitigating}, while projection introduces misalignment and occlusion-induced loss~\cite{Kweon2022Joint, odin}. These issues cause segmentation failures even for visually distinct regions, calling for a unified 3D representation that inherently encodes both geometry and appearance.

Among existing representations, 3D Gaussian Splatting (GS)~\cite{gaussiansplatting} offers a promising direction that can address both geometric sparsity and appearance deficiency. Unlike point clouds with binary occupancy, Gaussian primitives possess continuous volumetric attributes that encode both geometric structure and appearance properties. Residing in the same 3D coordinate frame as points, these attributes can transfer with less spatial misalignment than 2D-3D fusion~\cite{yu2024gaussian, Kweon2022Joint, mitigating, scenesplat}. However, optimization in GS often causes Gaussians to deviate from their initial positions~\cite{gaussiansplatting, scaffoldgs, guedon2024sugar, indoorgs}, creating a structural mismatch that prevents direct application of Gaussian attributes to point-level segmentation tasks. 

To address this challenge, we introduce Gaussian-to-Point (\textbf{G2P}), which augments input points with Gaussian attributes while preserving their original geometry. Since 3D GS primitives are optimized for photometric rendering and can deviate from the underlying surface geometry, directly using Gaussians as segmentation inputs may blur structures and break point-wise correspondence. We therefore keep the original point clouds (with accurate geometry) as input and use GS-derived attributes (\eg, opacity and scale) only as auxiliary cues (\cref{fig:1}(b)). This enables accurate segmentation of geometrically ambiguous objects that baseline methods tend to misclassify. Our G2P operates through three key components: (i)~\textit{Gaussian-to-Point Feature Augmentation} establishes precise correspondences via distance-based matching that accounts for anisotropic Gaussian ellipsoids; (ii)~\textit{GS Primitives-guided Learning} transfers view-consistent opacity cues to mitigate the point-level geometric bias; and (iii)~\textit{Scale-based Boundary Extraction} leverages anisotropic Gaussian scales to localize object boundaries for sharper segmentation.

We conduct comprehensive experiments on multiple benchmark datasets. Our G2P performs competitively with existing methods overall without direct 2D features or language supervision in the segmentation pipeline and shows consistent improvements on geometrically challenging classes. In summary, our main contributions are as follows:
\begin{itemize}
    \item We propose G2P, a covariance-aware Gaussian-to-Point alignment that transfers reliable 3D GS attributes (\eg, opacity and scale) to point clouds while preserving point geometry.
    \item We distill opacity-derived visibility cues from Gaussians into a point-only backbone, enabling Gaussian-free inference and bypassing cross-modal fusion.
    \item We improve boundary delineation by leveraging anisotropic Gaussian scales as a complementary boundary cue alongside semantic boundaries.
\end{itemize}

\section{Related Works}
\label{sec:related}

\subsection{3D Point Cloud Segmentation}
3D point cloud semantic segmentation has evolved from early per-point methods~\cite{pointcnn, pointnet, pointconv} to sophisticated architectures. PointNet++~\cite{pointnetplusplus} introduces hierarchical set abstraction for multi-scale learning, while MinkUNet~\cite{Minkwski} transforms point clouds into voxels and employs sparse convolutions to deliver robust and scalable segmentation. Recently, transformer-based architectures such as OctFormer~\cite{oct} and Point Transformer v3 (PT v3)~\cite{PointTFv3} have demonstrated state-of-the-art performance through efficient self-attention mechanisms.

Beyond supervised learning, self-supervised approaches like PointContrast~\cite{pointcontrast} leverage contrastive learning, while Sonata~\cite{sonata} demonstrates effective pre-training on large-scale 3D datasets. Despite these advances, purely geometric approaches suffer from boundary ambiguity and geometric bias in sparse point clouds~\cite{investigate, bgpseg}, particularly where appearance cues become critical for distinguishing geometrically similar objects.

\noindent\textbf{Boundary-Aware Approaches.}
Starting with~\cite{gong2021boundary}, recent boundary-aware frameworks such as JSENet~\cite{jsenet}, CBL~\cite{tang2022contrastive}, and BFANet~\cite{bfanet} explicitly learn boundary features to refine object boundaries in sparse point clouds~\cite{edgeaware}. These methods capture geometric discontinuities through dedicated boundary detection modules and edge-aware loss functions. However, they rely mainly on geometric cues and remain limited when segmentation requires appearance cues to address geometric bias arising from similar shapes but different materials.

\noindent\textbf{Cross-Modal 2D-3D Fusion.}
To address the appearance deficiency, cross-modal 2D-3D fusion approaches leverage rich visual features from RGB images. Multi-view methods such as VMVF~\cite{vmvf}, 3DMV~\cite{3dmv}, and MVPNet~\cite{jaritz2019multiview} aggregate features from multiple viewpoints through pooling or attention mechanisms. Subsequent studies~\cite{bpnet, robert2022learning, bridging, odin} further explore bidirectional projection and pre-trained backbone fusion. However, these approaches are hindered by point-to-pixel alignment errors in occluded regions, projection-induced information loss, and spatial misalignment between sparse 3D distributions and dense 2D grids~\cite{Kweon2022Joint, 3dpc-survey}. Consequently, 2D–3D fusion struggles to fully unify appearance and geometry within native 3D space~\cite{genova2021learning, interlaced}. Unlike prior 2D–3D fusion pipelines that rely on explicit 2D-to-3D projection, our approach explores an alternative direction by deriving 3D cues from GS-guided point attributes.

\subsection{Gaussian Splatting for 3D Segmentation}
With the advent of 3D Gaussian Splatting (GS)~\cite{gaussiansplatting}, recent works leverage GS for 3D segmentation and scene understanding~\cite{gaussiangrouping, flashsplat, segment3dgaussians}, and even for open-vocabulary or open-world perception~\cite{opengaussian, dcseg, drsplat, refersplat, reasongrounder, identitysplat, panogs, lu2025segment}. Beyond improvements in rendering quality, efficiency, and compression, Gaussians encode continuous volumetric geometry and appearance in 3D scenes. Building on this potential, numerous methods~\cite{feature3dgs, langsplat, trace3d, rethinking} lift CLIP~\cite{clip}-/SAM~\cite{sam}-derived features or masks into GS and optimize segmentation within the Gaussian representation. These methods focus primarily on view-consistent rendering with segmented Gaussians, which serves a different objective from point cloud segmentation. They also provide no explicit Gaussian-to-Point correspondences to preserve the original geometric structure, and instead serve as an intermediate representation for rendering purposes.

UniPre3D~\cite{unipre3d} targets point-level tasks, but relies on pixel-wise rendering losses combined with cross-modal fusion to pre-train a backbone network, treating Gaussians as an auxiliary self-supervised signal rather than producing supervised point-level outputs. By contrast, our G2P transfers GS attributes directly to points via explicit Gaussian-to-Point alignment under 3D supervision, without requiring recourse to 2D priors or rendering-based losses. Prior evidence shows that these Gaussian attributes encode both appearance and geometric properties~\cite{shapesplat, scenesplat, mitigating}. Building on this insight, G2P exploits them as complementary structural-confidence and boundary cues, coupling the view-consistent cues of Gaussians with the geometric stability of point clouds.

\begin{figure*}[t]
    \centering
    \includegraphics[width=\textwidth]{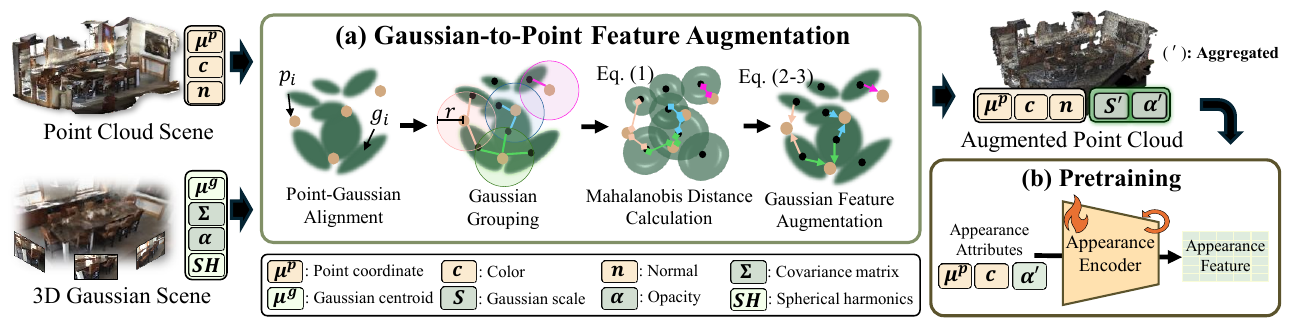}
    \caption{
        \textbf{Preparation stage of G2P.} 
In the preparation stage, (a) Gaussian-to-Point Feature Augmentation aligns 3D Gaussian (\( \mu^g, \Sigma, \alpha \)) and point (\( \mu^p, c, n \)), aggregating Gaussian attributes (\( \alpha, S \)), where \( S \) represents the scale derived from the \(\Sigma \), into augmented point features (\( \alpha', S' \)). (b) An appearance encoder is pre-trained in a self-supervised manner using the augmented point representation (\( \mu^p, c, \alpha' \)).}
    \label{fig:2}
\end{figure*}

\section{Method}
\subsection{Observation and Overview}

In real-world 3D environments, numerous challenging classes lack clear geometric discriminability, which makes segmentation difficult. For example, doors and windows that are coplanar with walls, and appliances with reflective surfaces positioned adjacent to walls, are often geometrically indistinct from background structures. As shown in~\cref{fig:1}(a), the baseline model fails to distinguish between the refrigerator and the wall, misclassifying them as a single planar surface. Since these objects are primarily distinguishable by appearance cues such as color and texture, which are easily recognizable by human vision, learning representations solely based on geometry produces geometrically biased features.

To overcome the geometric bias of conventional methods, we propose Gaussian-to-Point (G2P), a unified learning approach that enriches point cloud representations with 3D Gaussian attributes. G2P operates in two stages. In the preparation stage (\cref{fig:2,fig:3}), the Gaussian-to-Point Feature Augmentation module (\cref{subsec:G2P}) aligns Gaussians with points and augments Gaussian scale and opacity attributes while the Scale-based Boundary Extraction (\cref{subsec:bound}) derives boundary pseudo-labels from the aggregated scale distribution. With the augmented point clouds, we train an appearance encoder. In the training stage (\cref{fig:4}), GS Appearance Distillation (\cref{subsec:opacity}) employs the pre-trained appearance encoder to transfer features learned from point color and Gaussian opacity to the point cloud segmentation network. This design enriches the point cloud with Gaussian-derived opacity and scale cues while preserving its original point geometry.

\subsection{Preliminaries}
\label{subsec:pre}
3D GS is a view synthesis method that models a 3D scene as a set of anisotropic Gaussians and renders images through alpha blending. Each Gaussian is parameterized by centroid $\mu^g$, opacity $\alpha$, spherical harmonics $SH$, and a covariance matrix $\mathbf{\Sigma}$. The opacity $\alpha$ controls the blending contribution of each Gaussian along the ray, representing its visibility in the rendered view. The spatial shape of a Gaussian is defined by the covariance matrix $\mathbf{\Sigma} = \mathbf{R} \mathbf{S} \mathbf{S}^T \mathbf{R}^T$, where $\mathbf{R}$ is a rotation matrix and $\mathbf{S}$ is a diagonal scale matrix, with the scale denoted as $S$ throughout. The scale $S$ determines the directional spread and geometric extent of the Gaussian in 3D space. We use $S$ as a geometric cue and $\alpha$ as a view-consistent confidence cue from Gaussian primitives. We further analyze this interpretation in the Supplementary (Sec. B). These choices ensure that G2P focuses on informative attributes that are available for all scenes.

3D GS produces a set of optimized 3D Gaussians whose coordinates are altered during adaptive density optimization. As a result, the original geometric structure of the input point cloud is lost, and point-level semantic labels become unusable~\cite{gaussiansplatting, scaffoldgs, guedon2024sugar, indoorgs}. The resulting Gaussian scene exhibits substantially more noise and structural artifacts, such as indistinct edges, when compared to the cleaner input point cloud, making direct use for segmentation unreliable.

\subsection{Gaussian-to-Point Feature Augmentation}
\label{subsec:G2P}
To address these fundamental limitations, we propose the Gaussian-to-Point feature augmentation approach. This approach maintains the original point cloud geometry and instead aligns each point with its nearest 3D Gaussians. Each point is then augmented with the associated Gaussian attributes, enabling semantic segmentation while preserving geometric fidelity. As shown in~\cref{fig:2}(a), to augment the point cloud $\mathcal{P} = \{p_i\}_{i=1}^N$, where each point $p_i = (\mu_i^p, c_i, n_i) \in \mathbb{R}^{9}$ consists of coordinates, color, and a normal vector, we transfer attributes from a set of 3D Gaussians $\mathcal{G} = \{g_j\}_{j=1}^M$. Both sets are first aligned in the same coordinate space. For each point $p_i$, we employ a two-stage process to identify its $k$ most relevant Gaussian neighbors. First, to efficiently narrow down candidates, we select all Gaussians whose centroids fall within a Euclidean radius $r^g$ of $p_i$. Second, within this candidate group, we compute the Mahalanobis distance~\cite{Mahalanobis} to each Gaussian. Unlike the Euclidean distance that assumes isotropic distributions, it incorporates each Gaussian’s anisotropic shape determined by its scale $S$ and rotation $R$, providing a more physically plausible measure of proximity. The final $k$ neighbors are selected based on the smallest Mahalanobis distances. Accordingly, the proposed weight parameter $w_{ij}$ for each neighbor $g_j$ is determined by the inverse of its Mahalanobis distance, normalized across the $k$ neighbors:
\begin{equation}
w_{ij} = \frac{1 / \sqrt{(\mu_i^p - \mu_j^g)^T \Sigma_j^{-1} (\mu_i^p - \mu_j^g)}}{\sum_{l=1}^{k} \left( 1 / \sqrt{(\mu_i^p - \mu_l^g)^T \Sigma_l^{-1} (\mu_i^p - \mu_l^g)} \right)}.
\end{equation}
These weights are first used to aggregate the scale $S_j^g \in \mathbb{R}^{3}$ and opacity $\alpha_j^g \in \mathbb{R}$ attributes from the $k$ neighbors. This yields  the updated scale $S_i^p$ and weighted opacity $\alpha_i^p$ for each point $p_i$:
\begin{equation}
S_i^p = \sum_{j=1}^{k} w_{ij} \cdot S_j^g, \qquad \alpha_i^p = \sum_{j=1}^{k} w_{ij} \cdot \alpha_j^g.
\end{equation}
This augmentation extends each point $p_i$ to the form:
\begin{equation}
p_i = (\mu^p, c, n, S', \alpha') \in \mathbb{R}^{13}.
\end{equation}
As a result, the point cloud $\mathcal{P}$ is augmented with scale $S'$ and opacity $\alpha'$ attributes from the Gaussians. This attribute-augmented point cloud is further employed to train the appearance encoder and to derive scale-based boundary pseudo-labels. Algorithmic details of the G2P augmentation procedure are provided in the supplementary material (Alg.~1).

\begin{figure}[t]
  \centering
  \includegraphics[width=1.0\linewidth]{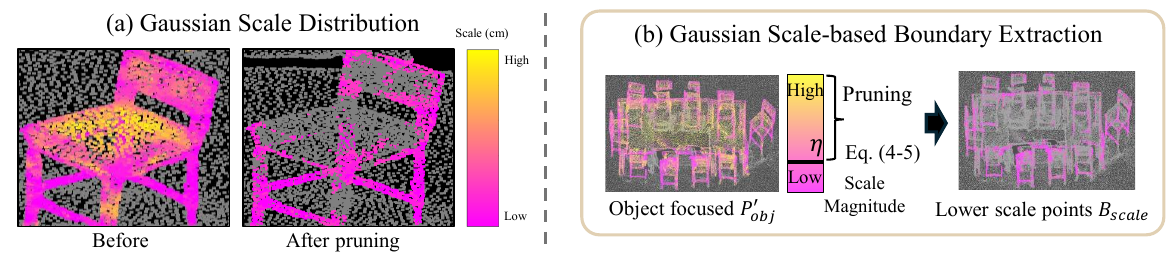}
  \caption{\textbf{Gaussian scale distribution and boundary extraction.} (a) Points after alignment, color-coded by aggregated Gaussian scale magnitude (magenta = small, yellow = large). Small scales concentrate at object boundaries. (b) Boundary pseudo-labels obtained by pruning large-scale points based on the scale criterion in \cref{eq:4,eq:5}.}
  \label{fig:3}
\end{figure}

\subsection{Gaussian Scale-based Boundary Extraction}
\label{subsec:bound}
Geometric discontinuities are often insufficient for boundaries in coplanar objects or thin structures. We instead exploit Gaussian scale attributes, where small aggregated scales tend to appear near object boundaries, while large scales dominate smooth planar regions (\cref{fig:3}).

Following attribute augmentation, we extract boundary pseudo-labels from the augmented point cloud $\mathcal{P}'$ using the acquired scale features $S_i'$. Specifically, we first form an object-focused point cloud $\mathcal{P}'_{\text{obj}}$ by removing points from background classes (\eg, floor, wall). For each point $p_i'$ with augmented scale vector $S_i' = (S_x', S_y', S_z')$, we compute its scale magnitude as the L2 norm:
\begin{equation}
\|S_i'\|_2 = \sqrt{S_x'^2 + S_y'^2 + S_z'^2}.
\label{eq:4}
\end{equation}
As illustrated in~\cref{fig:3}(b), we then prune points with large scale magnitudes by selecting a threshold $\tau_\eta$ that removes the top $100\eta\%$ of points with the largest scale magnitudes, and treat the remaining low-scale points as scale-based boundary candidates:
\begin{equation}
\mathcal{B}_{\text{scale}} = \left\{ p_i' \in \mathcal{P}'_{\text{obj}} \;\middle|\; \|S_i'\|_2 \le \tau_\eta \right\},
\label{eq:5}
\end{equation}
where $\tau_\eta$ is the scale-magnitude threshold determined by the pruning ratio $\eta$.

However, small scales can also arise from texture-induced or photometric variations, which may introduce noise when using $\mathcal{B}_{\text{scale}}$ alone. To complement this, we additionally derive semantic boundary candidates $\mathcal{B}_{\text{sem}}$: a point is marked as a semantic boundary if any neighboring point within a local radius $r^s$ has a different semantic label. Finally, we take the union of both cues to obtain the boundary pseudo-labels $\mathcal{B} = \mathcal{B}_{\text{scale}} \cup \mathcal{B}_{\text{sem}}$ for subsequent training.

\begin{figure*}[t]
    \centering
    \includegraphics[width=\textwidth]{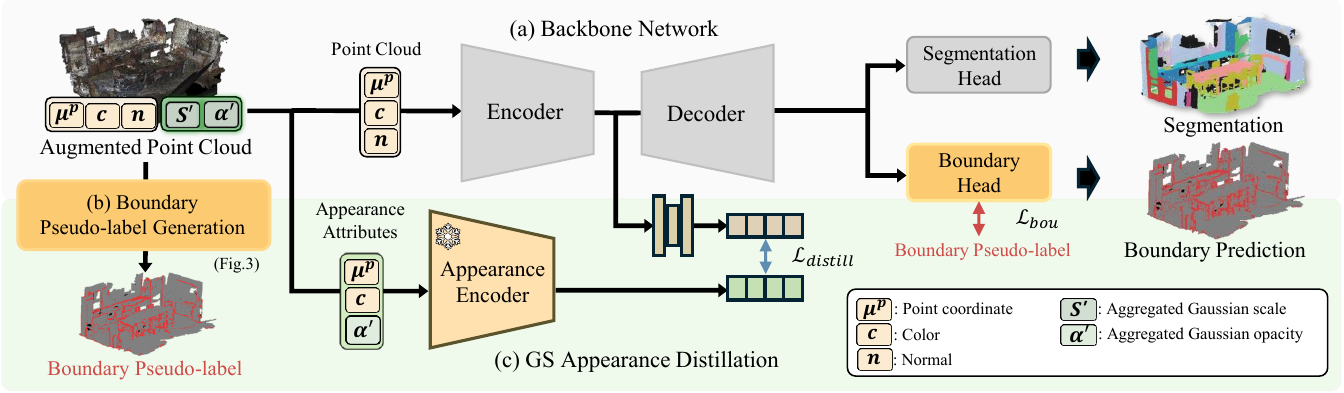}
    \caption{
        \textbf{Training stage of the G2P.} (a) In the training stage, the point cloud (\( \mu^p, c, n \)) is processed by the backbone network to produce semantic segmentation and boundary predictions. (b) Boundary pseudo-labels generated in the preparation stage supervise the boundary head, while (c) the GS appearance encoder provides appearance features that guide the backbone through appearance distillation. The model is trained with joint supervision from semantic segmentation, boundary prediction, and appearance distillation.
    }
    \label{fig:4}
\end{figure*}

\subsection{Gaussian Attribute-guided Learning}
\label{subsec:opacity}
\noindent\textbf{Appearance Encoder Pre-training.}
To preserve geometric consistency and prevent cross-modal misalignment, our approach takes point clouds as the sole input and receives auxiliary visual cues through feature distillation from a teacher encoder. To learn such appearance features, we repurpose the Sonata~\cite{sonata} architecture as the teacher encoder and train it \textbf{from scratch} on the augmented point representation. While Sonata is a self-supervised framework that independently trains an encoder, in our setup, we configure this framework to learn appearance features. As shown in~\cref{fig:2}(b), during the pre-training step, instead of using the traditional representation $p_i = (\mu^p, c, n)$ with 3D geometric normals $n \in \mathbb{R}^3$, we encode each point as $p_i = (\mu^p, c, \alpha')$, where $\alpha' \in \mathbb{R}$ denotes opacity, introducing view-consistent opacity cues in place of geometric information. The encoder is trained from scratch on the corresponding dataset (\eg, ScanNet v2~\cite{scannet} or ScanNet++~\cite{scannetpp}), ensuring dataset-specific pre-training for each benchmark. 

\noindent\textbf{Training Stage.}
As illustrated in~\cref{fig:4}, during training the PT v3~\cite{PointTFv3}-based backbone takes the input point cloud $p_i = (\mu^p, c, n)$ and jointly predicts semantic segmentation and boundary maps, where the boundary branch is supervised by pseudo-labels generated in the preparation stage. The pre-trained appearance encoder subsequently distills its representations into the backbone network. To incorporate boundary information into semantic segmentation, we adopt the boundary-semantic (B-S) block from BFANet~\cite{bfanet} and integrate it into the PT v3 backbone.

\noindent\textbf{Training Losses.}
Finally, the 3D segmentation backbone is trained using the proposed three supervisory signals: semantic supervision, boundary supervision, and appearance distillation. For the distillation signal, we employ an additional mapping MLP $\phi$, which takes the output features of the main point encoder, $f_i^p$, as input. We then distill knowledge from the features of the appearance encoder $f_i^a$ by minimizing the cosine similarity loss, $\mathcal{L}_{\text{distill}}$:
\begin{equation}
\mathcal{L}_{\text{distill}} = \frac{1}{N} \sum_{i=1}^{N} \left( 1 - \frac{\phi(f_i^p) \cdot f_i^a}{\|\phi(f_i^p)\|_2 \|f_i^a\|_2} \right).
\end{equation}
Semantic supervision loss, $\mathcal{L}_{\text{sem}}$, is applied via a combined loss of cross-entropy and Lovász-softmax~\cite{lova}, while boundary supervision loss, $\mathcal{L}_{\text{bou}}$, is provided using a combination of binary cross-entropy and Dice loss~\cite{diceloss}. For these losses, we denote the ground truth semantic labels as $y^g$, boundary pseudo-labels as $b^g$, and the corresponding model predictions as $\hat{p}$ and $\hat{b}$, respectively. Here, the loss functions are defined as:
\begin{equation}
\mathcal{L}_{\text{sem}} = \mathcal{L}_{\text{CE}}(y^g, \hat{p}) + \mathcal{L}_{\text{Lov}}(y^g, \hat{p}),
\end{equation}

\begin{equation}
\mathcal{L}_{\text{bou}} = \mathcal{L}_{\text{BCE}}(b^g, \hat{b}) + \mathcal{L}_{\text{Dice}}(b^g, \hat{b}).
\end{equation}
The final loss, $\mathcal{L}_{\text{total}}$, is a weighted sum of the three terms, where $\lambda_b$ and $\lambda_d$ are weights balancing the boundary and distillation losses, respectively:
\begin{equation}
\mathcal{L}_{\text{total}} = \mathcal{L}_{\text{sem}} + \lambda_b \mathcal{L}_{\text{bou}} + \lambda_d \mathcal{L}_{\text{distill}}.
\end{equation}

\section{Experiments}
\subsection{Experiment Setting}

\noindent\textbf{Dataset.}
We evaluate our approach on ScanNet v2~\cite{scannet}, ScanNet200~\cite{scannet200}, ScanNet++~\cite{scannetpp}, and Matterport3D~\cite{matterport3d}. 
These datasets contain 20, 200, 100, and 21 semantic classes, respectively. 
We report results on the validation splits for all datasets. For 3D Gaussian representations, we adopt the SceneSplat-7K~\cite{scenesplat} dataset, where each scene is reconstructed with approximately 1.5M Gaussian primitives. SceneSplat-7K provides GS reconstructions for indoor scenes, but outdoor environments are not available in the dataset.

\noindent\textbf{Evaluation Metric.}
Following prior works~\cite{PointTFv3, bfanet}, we adopt three standard metrics for evaluation: mean Intersection over Union (mIoU), overall Accuracy (OA), and mean Average Precision (mAP).

\begin{table}[t]
\centering

\begin{minipage}[c]{0.495\columnwidth}
\centering
\caption{\textbf{Semantic segmentation on the ScanNet v2 validation set.} $^*$ indicates external pre-training beyond ScanNet v2. $\dagger$ reproduced by us. \textbf{Bold} denotes the best in each column.}
\label{tab:1}
\setlength{\tabcolsep}{1pt}\footnotesize
\resizebox{\linewidth}{!}{
\begin{tabular}{lccc}
\toprule
\textbf{Method} & \textbf{Venue} & \textbf{Type} & \textbf{mIoU$\uparrow$} \\
\midrule
VMVF$^*$~\cite{vmvf} & ECCV'20 & \multirow{5}{*}{\shortstack{2D--3D \\ Fusion}} & 76.4 \\
BPNet~\cite{bpnet} & CVPR'21 &  & 69.7 \\
ODIN$^*$ (Swin-B)~\cite{odin} & CVPR'24 &  & 77.8 \\
PonderV2$^*$~\cite{ponderv2} & TPAMI'25 &  & 77.0 \\
UniPre3D~\cite{unipre3d} & CVPR'25 &  & 77.6 \\
\cmidrule(lr){1-4}
MinkUNet~\cite{Minkwski} & CVPR'19 &  & 72.2 \\
OctFormer~\cite{oct} & TOG'23 & & 75.7 \\
SPG~\cite{han2024subspace} & ECCV'24 & \multirow{2}{*}{Geometric} & 76.0 \\
PT v3~\cite{PointTFv3} & CVPR'24 &  & 77.5 \\
\textcolor{gray!50}{PT v3 + PPT$^*$~\cite{PointTFv3}} & \textcolor{gray!50}{CVPR'24} &  & \textcolor{gray!50}{78.6} \\
BFANet$^\dagger$~\cite{bfanet} & CVPR'25 &  & 77.3 \\
BFANet~\cite{bfanet} & CVPR'25 &  & 78.0 \\
\midrule
\rowcolor{gray!12}\textbf{G2P (Ours)} & -- & \textbf{GS-guided} & \textbf{78.4} \\
\bottomrule
\end{tabular}
}
\end{minipage}
\hfill
\begin{minipage}[c]{0.495\columnwidth}
\centering
\caption{\textbf{Semantic segmentation on the ScanNet200 validation set.} $^*$ indicates external pre-training beyond ScanNet v2. 
\textbf{Bold} denotes the best in each column.}
\label{tab:2}
\setlength{\tabcolsep}{2pt}\footnotesize
\renewcommand{\arraystretch}{0.9}
\resizebox{0.9\linewidth}{!}{
\begin{tabular}{lccc}
\toprule
\textbf{Method} & \textbf{Venue} & \textbf{mIoU$\uparrow$} & \textbf{OA$\uparrow$} \\
\midrule
MinkUNet~\cite{Minkwski} & CVPR'19 & 25.0 & 80.4 \\
PointContrast~\cite{pointcontrast} & ECCV'20 & 26.2 & -- \\
PT v2~\cite{wu2022point} & NeurIPS'22 & 30.2 & 82.7 \\
OctFormer~\cite{oct} & TOG'23 & 32.6 & 83.0 \\
SPG~\cite{han2024subspace} & ECCV'24 & 31.5 & -- \\
PT v3~\cite{PointTFv3} &
CVPR'24&
35.2 &
83.6 \\
\textcolor{gray!50}{PT v3 + PPT$^*$~\cite{PointTFv3}} &
\textcolor{gray!50}{CVPR'24} &
\textcolor{gray!50}{36.0} &
-- \\
PonderV2$^*$~\cite{ponderv2} & TPAMI'25 & 32.3 & -- \\
UniPre3D$^*$~\cite{unipre3d} & CVPR'25 & 36.0 & 83.7 \\
\midrule
\rowcolor{gray!12}
\textbf{G2P (Ours)} & -- & \textbf{36.6} & \textbf{83.8} \\
\bottomrule
\end{tabular}
}
\end{minipage}
\end{table}

\noindent\textbf{Implementation Details.}
Our main training is conducted on a single NVIDIA RTX 3090 GPU for 800 epochs. The appearance encoder pre-training, following Sonata~\cite{sonata}, is performed separately on an NVIDIA A6000 GPU for 400 epochs per dataset from scratch. We adopt PT v3~\cite{PointTFv3} as the backbone and incorporate the B-S block from BFANet~\cite{bfanet}. The training batch size is set to 4, and we adopt AdamW~\cite{adamw} as the optimizer with an initial learning rate of 0.003. All other settings follow PT v3, and detailed architectural configurations are reported in the Supplementary (Sec. A). For Gaussian-to-Point augmentation, we set the Gaussian candidate search radius to $r^g = 0.06$ m and use $k=20$ Gaussian neighbors. For boundary pseudo-label extraction, we set the scale-based trimming ratio $\eta=0.7$. The radius $r^s$ for semantic boundary calculation is set to 0.04 m. Loss weights are balanced as $\lambda_d$ = 0.4, $\lambda_b$ = 0.9.

\subsection{Quantitative Comparison}
\noindent\textbf{Results on ScanNet v2.}
We compare G2P with geometric methods and 2D-3D fusion approaches on the ScanNet v2 validation set. \cref{tab:1} reports the results. G2P exceeds PT v3 by \textbf{+0.9} mIoU and BFANet~\cite{bfanet} by \textbf{+0.4}. G2P outperforms all geometric and 2D–3D fusion baselines without external pre-training. For reference, ODIN~\cite{odin}, PonderV2~\cite{ponderv2}, and PT v3 + PPT~\cite{PointTFv3} use external pre-training beyond ScanNet v2, yet G2P achieves competitive performance compared to PT v3 + PPT without such pre-training.

\noindent\textbf{Results on ScanNet200.}
We further evaluate on the ScanNet200~\cite{scannet200} validation set, which expands the label space from 20 to 200 fine-grained categories (\cref{tab:2}). G2P attains \textbf{36.6} mIoU and \textbf{83.8} OA, improving over PT v3~\cite{PointTFv3} baseline by \textbf{+1.4} mIoU and remaining competitive with UniPre3D~\cite{unipre3d}. ScanNet200 contains many fine-grained categories with subtle geometric differences, where GS-derived cues become particularly beneficial. 

\begin{table*}[t]
\centering
\caption{\textbf{Class-wise IoU comparison on ScanNet~v2 categories.}
\textcolor{red}{\textbf{Red}} and \textcolor{blue}{\underline{blue}} denote the best and second-best IoU, respectively. $\dagger$ reproduced by us. \textbf{Avg.}: Average IoU for each subset. \textbf{Refrig.}: Refrigerator. \textbf{ShwrCurt.}: Shower Curtain.}
\label{tab:3}
\setlength{\tabcolsep}{3.5pt}\footnotesize
\resizebox{\textwidth}{!}{
\begin{tabular}{c|cccccccc|c|cccccc|c}
\toprule
\multirow{2}{*}{\textbf{Method}} &
\multicolumn{9}{c|}{\textbf{Geometrically Distinguishable Classes}} &
\multicolumn{7}{c}{\textbf{Geometrically Challenging Classes}} \\
\cmidrule(lr){2-10} \cmidrule(lr){11-17}
& \textbf{Wall} & \textbf{Floor} & \textbf{Cabinet} & \textbf{Bed} & \textbf{Chair} & \textbf{Sofa} & \textbf{Table} & \textbf{Bookshelf} & \textbf{Avg.} &
\textbf{Door} & \textbf{Window} & \textbf{Picture} & \textbf{Curtain} & \textbf{Refrig.} & \textbf{ShwrCurt.} & \textbf{Avg.} \\
\midrule
MinkUNet$^\dagger$~\cite{Minkwski} &
87.1 & \textcolor{red}{\textbf{96.7}} & 67.0 & 83.4 & 92.6 & 84.5 & 77.0 & 81.4 & 83.7 &
70.4 & 65.0 & 37.8 & 77.2 & 64.0 & 67.2 & 63.6 \\
OctFormer$^\dagger$~\cite{oct} &
86.1 & 95.7 & 70.1 & 82.7 & 91.9 & 83.6 & 74.1 & 81.3 & 83.2 &
67.7 & 67.2 & 34.0 & 77.4 & 65.5 & 65.8 & 62.9 \\
PT v3$^\dagger$~\cite{PointTFv3} &
\textcolor{red}{\textbf{87.7}} & \underline{\textcolor{blue}{95.8}} & \underline{\textcolor{blue}{73.0}} & 83.2 & \textcolor{red}{\textbf{93.1}} & 83.7 & 78.9 & 80.6 & 84.5 &
\underline{\textcolor{blue}{74.6}} & 72.8 & \textcolor{red}{\textbf{41.8}} & 77.9 & 64.9 & 68.9 & 66.8 \\
UniPre3D~\cite{unipre3d} &
87.5 & 95.5 & 72.6 & 84.4 & 92.7 & \textcolor{red}{\textbf{84.9}} & \underline{\textcolor{blue}{80.0}} & \underline{\textcolor{blue}{82.3}} & \underline{\textcolor{blue}{85.0}} &
\underline{\textcolor{blue}{74.6}} & 72.6 & 38.5 & \textcolor{red}{\textbf{79.0}} & \underline{\textcolor{blue}{70.3}} & \textcolor{red}{\textbf{76.3}} & \underline{\textcolor{blue}{68.6}} \\
BFANet$^\dagger$~\cite{bfanet} &
87.2 & 95.7 & 72.3 & \underline{\textcolor{blue}{84.7}} & 92.5 & 83.0 & \textcolor{red}{\textbf{81.2}} & 80.6 & 84.7 &
74.0 & \underline{\textcolor{blue}{73.5}} & 38.5 & 77.5 & 69.3 & 70.8 & 67.3 \\
\midrule
\rowcolor{gray!12}\textbf{G2P (Ours)} &
\underline{\textcolor{blue}{87.6}} & \underline{\textcolor{blue}{95.8}} & \textcolor{red}{\textbf{73.3}} & \textcolor{red}{\textbf{87.8}} & \underline{\textcolor{blue}{93.0}} & \underline{\textcolor{blue}{84.6}} & 79.7 & \textcolor{red}{\textbf{84.6}} & \textcolor{red}{\textbf{85.8}} &
\textcolor{red}{\textbf{76.8}} & \textcolor{red}{\textbf{73.9}} & \underline{\textcolor{blue}{39.4}} & \underline{\textcolor{blue}{78.0}} & \textcolor{red}{\textbf{70.9}} & \underline{\textcolor{blue}{76.1}} & \textcolor{red}{\textbf{69.2}} \\
\bottomrule
\end{tabular}
}
\end{table*}

\noindent\textbf{Class-wise Analysis.}
To better understand where G2P excels, we categorize the 20 classes in ScanNet v2 into two groups based on geometric discriminability. \textit{Geometrically challenging classes} include objects that are difficult to distinguish by geometry alone and are often coplanar with walls or exhibit reflective surfaces. \textit{Geometrically distinguishable classes} include objects with clear structural features. As shown in \cref{tab:3}, our G2P achieves higher average IoU on geometrically challenging classes, reaching 69.2. The largest class-level improvements occur on \textit{refrigerator} (70.9 IoU, +6.0 vs. PT v3's 64.9) and \textit{shower curtain} (+7.2 IoU).

\begin{table}[t]
\centering

\begin{minipage}[c]{0.475\columnwidth}
\centering
\caption{\textbf{Evaluation on ScanNet200 Hidden Test Set.}
We report mIoU, Head, Common, and Tail IoU on the official ScanNet200 benchmark.}
\label{tab:4}
\footnotesize
\setlength{\tabcolsep}{2.0pt}
\renewcommand{\arraystretch}{1.0}
\resizebox{\linewidth}{!}{
\begin{tabular}{lcccc}
\toprule
\textbf{Method} & \textbf{mIoU} & \textbf{Head} & \textbf{Common} & \textbf{Tail} \\
\midrule
MinkUNet~\cite{Minkwski} & 25.3 & 46.3 & 15.4 & 10.6 \\
OctFormer~\cite{oct} & 32.6 & 53.9 & 26.5 & 13.1 \\
CeCo~\cite{ceco} & 34.0 & 55.1 & 24.7 & 18.1 \\
PonderV2~\cite{ponderv2} & 34.6 & \underline{\textcolor{blue}{55.2}} & 27.0 & 17.5 \\
BFANet~\cite{bfanet} & \textcolor{red}{\textbf{36.0}} & \textcolor{red}{\textbf{55.3}} & \textcolor{red}{\textbf{29.3}} & \underline{\textcolor{blue}{19.3}} \\
\midrule
\rowcolor{gray!12}
\textbf{G2P (Ours)} & \underline{\textcolor{blue}{35.7}} & 54.3 & \underline{\textcolor{blue}{28.7}} & \textcolor{red}{\textbf{20.2}} \\
\bottomrule
\end{tabular}
}
\end{minipage}
\hspace{0.01\columnwidth}%
\begin{minipage}[c]{0.505\columnwidth}
\centering
\caption{\textbf{Instance segmentation on ScanNet200 validation set.}
All methods use PointGroup~\cite{jiang2020pointgroup} as the instance segmentation framework, varying only the backbone.}
\label{tab:5}
\footnotesize
\setlength{\tabcolsep}{3.2pt}
\renewcommand{\arraystretch}{1.05}
\resizebox{0.80\linewidth}{!}{
\begin{tabular}{l|ccc}
\toprule
\textbf{Method} & mAP$_{25}$ & mAP$_{50}$ & mAP \\
\midrule
MinkUNet~\cite{Minkwski} & 32.2 & 24.5 & 15.8 \\
PT v2~\cite{wu2022point} & 39.6 & 31.9 & 21.4 \\
PT v3~\cite{PointTFv3} & 40.1 & 33.2 & 23.1 \\
\midrule
\rowcolor{gray!12}
\textbf{G2P (Ours)} & \textbf{41.8} & \textbf{33.3} & \textbf{23.2} \\
\bottomrule
\end{tabular}
}
\end{minipage}
\end{table}

We further evaluate our method on the ScanNet200 hidden test set (\cref{tab:4}). Following BFANet, we compare with methods that do not use large-scale pre-training beyond ScanNet v2. G2P achieves competitive performance with 35.7 mIoU, closely approaching BFANet's 36.0 while achieving the best performance on Tail classes.

\begin{figure*}[!t]
    \centering    
    \includegraphics[width=\textwidth]{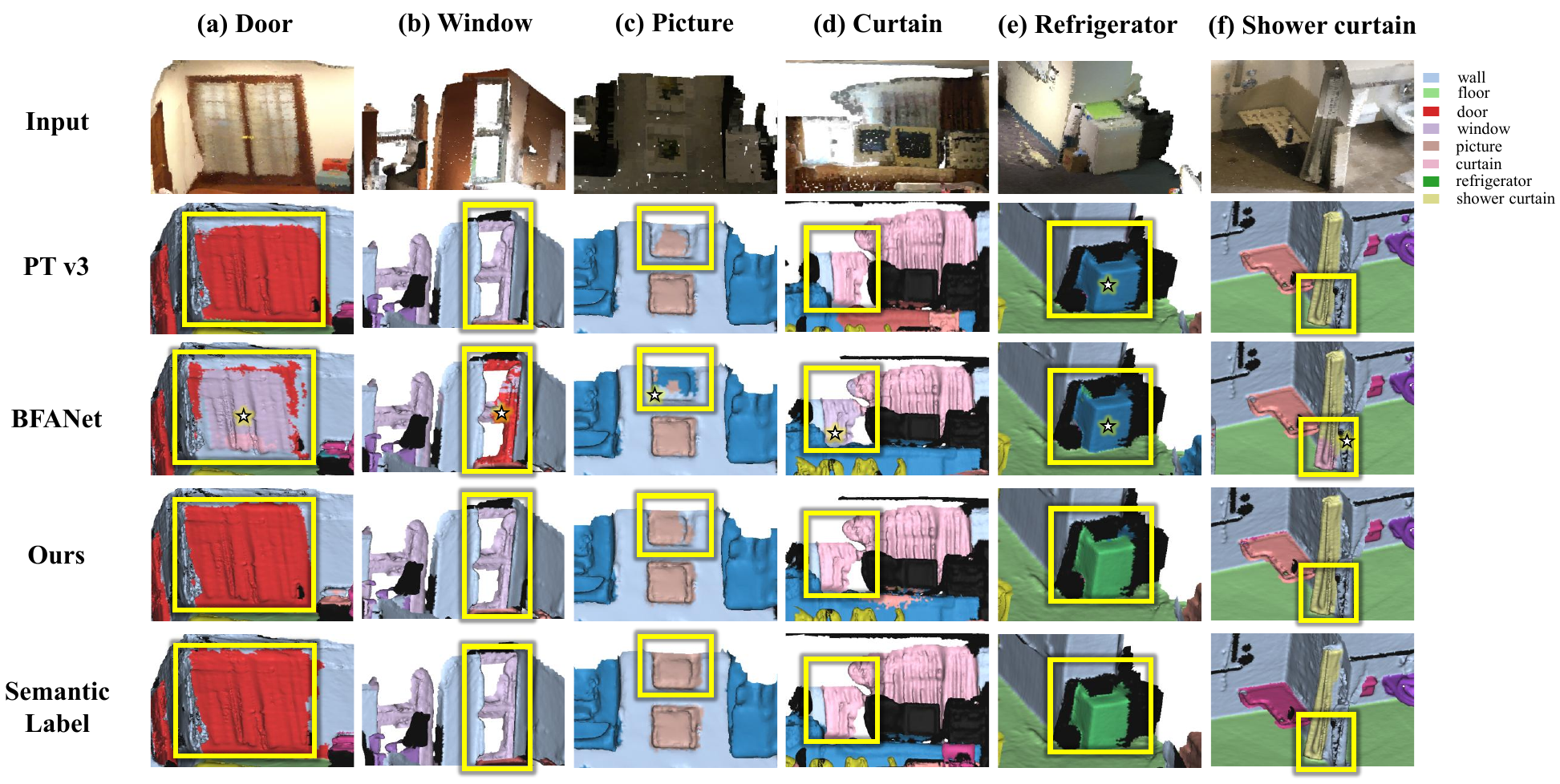}
\caption{\textbf{Qualitative comparison on geometrically challenging classes in ScanNet v2.} 
\textcolor{hiYellow}{\textbf{Yellow}} boxes highlight regions where models differ. Within these boxes, a \textbf{star (\textcolor{hiYellow}{\textbf{$\star$}})} denotes \textbf{category confusion}, where a baseline assigns a label inconsistent with the semantic label (\eg, door to wall, window to door, refrigerator to cabinet) due to geometric bias, whereas boxes without a star denote incomplete segmentation, where baselines fail to cover the full extent of thin or coplanar structures. Corresponding regions in the semantic label (bottom row) are marked for reference.}
    \label{fig:5}
\end{figure*}

\noindent\textbf{Instance Segmentation Results.}
We evaluate G2P on the ScanNet200 instance segmentation benchmark and report mAP, mAP$_{25}$, and mAP$_{50}$ on the validation set (\cref{tab:5}). Following standard practice, we adopt PointGroup~\cite{jiang2020pointgroup} as a unified instance segmentation framework. G2P attains 41.8 mAP$_{25}$, 33.3 mAP$_{50}$, and 23.2 mAP, improving over PT v3~\cite{PointTFv3} baseline by +1.7 in mAP$_{25}$. These results indicate that the proposed Gaussian-guided representation also benefits instance-level grouping. 

\subsection{Qualitative Comparisons}
\noindent\textbf{Qualitative Results on Challenging Classes.}
\cref{fig:5} compares G2P with PT v3~\cite{PointTFv3} and BFANet~\cite{bfanet}. Baselines show two failure modes, category confusion (\textcolor{hiYellow}{\textbf{$\star$}}) and incomplete segmentation, across challenging classes such as (a), (b), and (e), and thin or coplanar structures like (c), (d), and (f). G2P produces label-consistent and more complete masks in both cases. This aligns with \cref{tab:3}, where G2P achieves +6.0 IoU on \textit{refrigerator} over PT v3. Additional qualitative results and boundary predictions on ScanNet v2 are provided in the Supplementary.

\subsection{Ablation Studies}
\noindent\textbf{Results on Additional Benchmarks.}
To further validate G2P's effectiveness, we evaluate on the ScanNet++~\cite{scannetpp} and Matterport3D~\cite{matterport3d} validation sets. As shown in \cref{tab:6}, G2P generalizes well across datasets, achieving 48.7 and 55.9 mIoU on ScanNet++ and Matterport3D. Qualitative comparisons on these datasets are provided in the Supplementary (Sec. D).

\begin{table}[t]
\centering

\begin{minipage}[t]{0.475\columnwidth}
\centering
\caption{\textbf{Evaluation on ScanNet++ and Matterport3D validation set.}
\textbf{Bold} indicates the best in each column.
$\dagger$ denotes reproduced results.}
\label{tab:6}
\footnotesize
\setlength{\tabcolsep}{3pt}
\resizebox{0.9\linewidth}{!}{
\begin{tabular}{l|c|c}
\toprule
\textbf{Method} & \textbf{ScanNet++} & \textbf{Matterport3D} \\
 & \textbf{mIoU$\uparrow$} & \textbf{mIoU$\uparrow$} \\
\midrule
MinkUNet$^\dagger$~\cite{Minkwski} & 28.8 & 54.2 \\
OctFormer$^\dagger$~\cite{oct}     & 44.5 & 55.0 \\
PT v3$^\dagger$~\cite{PointTFv3}   & 47.9 & 55.5 \\
\midrule
\rowcolor{gray!12}
\textbf{G2P (Ours)}                & \textbf{48.7} & \textbf{55.9} \\
\bottomrule
\end{tabular}
}
\end{minipage}
\hspace{0.02\columnwidth}
\begin{minipage}[t]{0.475\columnwidth}
\centering
\caption{\textbf{Module ablation on ScanNet v2.}
Boundary denotes the boundary-guidance branch with boundary pseudo-label supervision.
Distillation denotes GS-attribute distillation.}
\label{tab:7}
\footnotesize
\setlength{\tabcolsep}{4pt}
\resizebox{0.95\linewidth}{!}{
\begin{tabular}{lcc|c}
\toprule
\textbf{Setting} & \textbf{Boundary} & \textbf{Distill.} & \textbf{mIoU$\uparrow$} \\
\midrule
Baseline (PT v3$^\dagger$~\cite{PointTFv3}) &  &  & 77.0 \\
Boundary only & \checkmark &  & 77.8 \\
Distillation only &  & \checkmark & 77.8 \\
\midrule
\rowcolor{gray!12}
\textbf{G2P (Full)} & \checkmark & \checkmark & \textbf{78.4} \\
\bottomrule
\end{tabular}
}
\end{minipage}

\end{table}

\noindent\textbf{Module Ablation.}
We conduct ablation studies to validate the contribution of each component (\cref{tab:7}). Both boundary guidance and GS-attribute distillation independently improve mIoU, with their combination achieving the highest gain. These results suggest that boundary guidance and GS-attribute distillation provide complementary benefits. Each component alone produces competitive results, and combining them yields reliable segmentation by reducing boundary leakage and false merges.

\begin{table}[t]
\centering

\begin{minipage}[c]{0.48\columnwidth}
\centering
\caption{\textbf{Ablation on boundary pseudo-label generation.}
The table compares semantic-based ($\mathcal{B}_{\text{sem}}$), only scale-based ($\mathcal{B}_{\text{scale}}$), and combined boundary pseudo-labels. Boundary guidance is evaluated without distillation and with the boundary loss fixed at $\lambda_b = 1.0$. Results on ScanNet v2 validation set.}
\label{tab:8}
\footnotesize
\setlength{\tabcolsep}{4pt}
\resizebox{0.75\linewidth}{!}{
\begin{tabular}{cc|c}
\toprule
$\mathcal{B}_{\text{sem}}$ (radius $r^s$) & $\mathcal{B}_{\text{scale}}$ (ratio $\eta$) & \textbf{mIoU$\uparrow$} \\
\midrule
$r^s=0.02$ & -- & 76.5 \\
$r^s=0.04$ & -- & 76.9 \\
$r^s=0.06$ & -- & 76.3 \\
\midrule
-- & $\eta=0.3$ & 77.3 \\
-- & $\eta=0.5$ & 76.5 \\
-- & $\eta=0.7$ & 76.9 \\
\midrule
$r^s=0.02$ & $\eta=0.7$ & 77.6 \\
\rowcolor{gray!12}
$r^s=0.04$ & $\eta=0.7$ & \textbf{77.8} \\
$r^s=0.06$ & $\eta=0.7$ & 76.9 \\
\bottomrule
\end{tabular}
}

\end{minipage}
\hspace{0.02\columnwidth}
\begin{minipage}[c]{0.48\columnwidth}
\centering
\caption{\textbf{Ablation on Gaussian primitive learning approach.}
Here, $c'$ and $n'$ denote Gaussian color and normal features aggregated from neighboring points based on Euclidean distance.
AE denotes the appearance encoder. AE fine-tuning follows the Sonata pipeline~\cite{sonata}; in Distillation, the listed attributes are used for the AE teacher.}
\label{tab:9}
\footnotesize
\setlength{\tabcolsep}{4pt}
\resizebox{\linewidth}{!}{
\begin{tabular}{lcc}
\toprule
\textbf{Strategy} & \textbf{Input Attributes} & \textbf{mIoU$\uparrow$} \\
\midrule

Raw 3D Gaussians 
& $\mu^g, c', n', \alpha$ 
& 73.1 \\

Augmented points 
& $\mu^p, c, n, \alpha'$ 
& 77.3 \\

\midrule

AE fine-tuning
& $\mu^p, c, \alpha'$ 
& 74.9 \\

AE fine-tuning
& $\mu^p, c, n, \alpha'$ 
& 75.0 \\

\midrule

\rowcolor{gray!12}
Distillation (ours) 
& $\mu^p, c, \alpha'$ 
& \textbf{77.8} \\

\bottomrule
\end{tabular}
}
\end{minipage}

\end{table}

\noindent\textbf{Ablations on Boundary Pseudo-label Generation.}
\cref{tab:8} compares different boundary pseudo-label formulations. The semantic-based variant ($\mathcal{B}_{\text{sem}}$), commonly adopted in prior work~\cite{bpnet,bfanet}, provides the baseline boundary supervision. The scale-based version ($\mathcal{B}_{\text{scale}}$) shows comparable performance to the semantic-based approach. The combined formulation ($\mathcal{B} = \mathcal{B}_{\text{sem}} \cup \mathcal{B}_{\text{scale}}$) yields the highest mIoU among the tested configurations. Variations in the semantic radius $r^s$ indicate modest sensitivity.

\noindent\textbf{Ablation on Gaussian Attribute Learning.}
\cref{tab:9} evaluates strategies for leveraging GS-derived attributes. Using raw Gaussian coordinates proves unreliable for point-level segmentation (73.1 mIoU). In contrast, \textit{Augmented points} improves performance to 77.3 mIoU, demonstrating that aggregated Gaussian opacity provides complementary cues. Notably, directly fine-tuning the appearance encoder with these opacity cues yields only 74.9–75.0 mIoU, indicating that distillation is a more effective mechanism for transferring GS-guided representations to a point-only backbone. Our distilled variant reaches the highest mIoU of 77.8 and removes the need for Gaussian features at inference.

\begin{table}[t]
\centering
\caption{\textbf{Ablation on distance metric and neighborhood size $k$ (ScanNet v2).} $k$ denotes the neighborhood size used for Mahalanobis distance.} 
\label{tab:10}
\scriptsize
\setlength{\tabcolsep}{3pt}
\renewcommand{\arraystretch}{1.0}
\begin{tabular}{ccccc}
\toprule
\textbf{Metric} & Euclidean & $k=10$ & $k=20$ & $k=30$ \\
\midrule
mIoU$\uparrow$ & 77.0 & 77.4 & \textbf{78.4} & 77.9 \\
\bottomrule
\end{tabular}
\end{table}
\begin{table}[!t]
\centering
\caption{\textbf{Model efficiency comparison on ScanNet v2.} $\dagger$ denotes reproduced results.}
\label{tab:11}
\scriptsize
\setlength{\tabcolsep}{2pt}

\resizebox{0.85\linewidth}{!}{
\begin{tabular}{l ccc|cc|c}
\toprule
\multirow{2}{*}{\textbf{Method}} & \multirow{2}{*}{\textbf{Params.}} &
\multicolumn{2}{c|}{\textbf{Training}} &
\multicolumn{2}{c|}{\textbf{Inference}} &
\multirow{2}{*}{\textbf{mIoU$\uparrow$}} \\
\cmidrule(lr){3-4}\cmidrule(lr){5-6}
 &  & Latency & Memory & Latency & Memory &  \\
\midrule
MinkUNet$^\dagger$~\cite{Minkwski} & 39.2M & 71ms & 1.6G & 29ms & 1.4G & 72.3 \\
OctFormer$^\dagger$~\cite{oct} & 44.0M & 259ms & 4.2G & 94ms & 4.4G & 74.3 \\
PT v3$^\dagger$~\cite{PointTFv3} & 46.2M & 132ms & 5.6G & 79ms & 1.9G & 77.0 \\
\rowcolor{gray!10}
\textbf{G2P (Ours)} & 46.4M & 220ms & 7.3G & 96ms & 3.3G & \textbf{78.4} \\
\bottomrule
\end{tabular}
}
\end{table}

\noindent\textbf{Correspondence Metric.}
Gaussian-to-Point correspondence depends on the distance metric and the neighborhood size $k$. Euclidean distance ignores the anisotropy of splats and may lead to unreliable matches near object boundaries. Mahalanobis~\cite{Mahalanobis} distance leverages the covariance $\boldsymbol{\Sigma}$ to form tighter, more reliable matches. As shown in \cref{tab:10}, a moderate neighborhood ($k=20$) offers the best trade-off (78.4 mIoU). Smaller $k$ is noisy, whereas larger $k$ oversmooths labels. These results support covariance-aware (Mahalanobis) matching as the default in G2P, confirming the benefit of exploiting Gaussian anisotropy.

\noindent\textbf{Inference Efficiency.}
\cref{tab:11} evaluates computational efficiency. Training G2P incurs additional overhead due to appearance distillation, with training latency increasing by +67\% and memory by +30\%. However, inference overhead remains moderate: parameters increase by only 0.2M (+0.4\%), latency by +21\%, and memory by +74\%. 

\section{Conclusion}
We present G2P, a novel approach for point cloud segmentation that integrates 3D GS to unify geometric and visibility cues in 3D space. Our three-component approach, comprising Gaussian-to-Point feature augmentation, opacity-guided GS representation learning, and scale-based boundary extraction, helps mitigate geometric bias. By augmenting point features with Gaussian attributes, G2P enhances segmentation on point clouds, particularly for objects that are geometrically ambiguous but have distinctive appearances. Extensive experiments on standard benchmarks demonstrate competitive performance and consistent gains across datasets. Furthermore, we conduct ablations to validate the effectiveness of GS-guided feature augmentation, establishing its practical value for 3D scene understanding. A current limitation is that G2P relies on an offline preparation stage and thus depends on the availability of input Gaussians. Extending it to outdoor scenes may require more robust GS construction.

\section*{Acknowledgements}
This work was supported in part by the National Research Foundation of Korea (NRF) grant funded by the Korea government (MSIT) (RS-2025-00520308) and in part by the Nurturing Global Technical Experts for Performance Evaluation of Multimodal Content Copyright Core Technologies Project through the Korea Creative Content Agency (KOCCA), funded by the Ministry of Culture, Sports and Tourism (MCST) and the Korea Creative Content Agency for Culture Technology (KCTIEP) (Project No.~RS-2026-2552393).

\title{G2P: Gaussian-to-Point Attribute Alignment for Boundary-Aware 3D Segmentation \\
\large Supplementary Materials}

\titlerunning{G2P: Gaussian-to-Point Attribute Alignment}

\author{%
Hojun Song\inst{1}$^{*}$
\and
Chae-yeong Song\inst{1,2}$^{*,\ddagger}$
\and
Jeong-hun Hong\inst{1}
\and
Chaewon Moon\inst{1}
\and \\
Soo Ye Kim\inst{3}
\and
Yiyi Liao\inst{4}
\and
Jaehyup Lee\inst{1}
\and
Sang-hyo Park\inst{1}$^{\dagger}$
}
\authorrunning{H.~Song et al.}
\institute{%
Kyungpook National University
\and
Korea Electronics Technology Institute
\and
Adobe Research
\and
Zhejiang University
}

\markboth{H.~Song et al.}{G2P: Gaussian-to-Point Attribute Alignment}

\appendix
\setcounter{section}{0}
\renewcommand{\thesection}{\Alph{section}}
\maketitle
\renewcommand{\thefootnote}{}
\footnotetext{$^{*}$~Equal contribution.\quad$^{\dagger}$~Corresponding author.}
\footnotetext{$^{\ddagger}$~Work done while at Kyungpook National University.}
\renewcommand{\thefootnote}{\arabic{footnote}}

\setcounter{figure}{5}
\setcounter{table}{11}

\setcounter{page}{1}

This supplementary material provides additional details and analyses that complement the main paper. Specifically, it includes architectural and implementation details (\cref{sec:Imple}), further analysis on how Gaussian scale and opacity guide segmentation (\cref{sec:analysis}), ablation studies validating each design choice (\cref{sec:ablation}), 
and additional qualitative visualizations (\cref{sec:add_Vis}). Discussion on limitations is also included (\cref{sec:limitation}). 

\section{Implementation Details and Experimental Protocols}
\label{sec:Imple}

\subsubsection{Architectural configurations.} We use the Sonata framework~\cite{sonata} to pre-train our appearance encoder on the Gaussian-to-Point augmented input, where each point is represented as $(\mu^{p}, c, \alpha') \in \mathbb{R}^{7}$. It is trained for 400 epochs with a batch size of 1 and a learning rate of 0.002; all other settings follow Sonata.

The main segmentation backbone is PT v3~\cite{PointTFv3}, and we insert a B--S block following BFANet~\cite{bfanet}, which takes 64-channel PT v3 decoder features and produces 128-dimensional boundary-aware features with 8 attention heads, following BFANet without modification. The boundary prediction head operates on the fused 128-dimensional feature and uses a two-layer MLP with a 64-dimensional hidden layer and a 1-dimensional output layer. All other settings follow PT v3.

\begin{table}[h]
\centering
\caption{\textbf{Training and evaluation protocol.}
For each benchmark, the appearance encoder is pre-trained only on the training split, and the segmentation model is trained and evaluated using the corresponding train/validation split.}
\label{tab:12}
\footnotesize
\setlength{\tabcolsep}{3.5pt}
\renewcommand{\arraystretch}{0.95}
\begin{tabular}{lccc}
\toprule
\textbf{Dataset} & \textbf{Appearance Encoder} & \textbf{Train} & \textbf{Evaluation} \\
\midrule
ScanNet v2~\cite{scannet}   & train only & train & val. \\
ScanNet++~\cite{scannetpp}    & train only & train & val. \\
Matterport3D~\cite{matterport3d} & train only & train & val. \\
\bottomrule
\end{tabular}
\end{table}

\subsubsection{Training and evaluation protocol.}
To clarify the training and evaluation setup, we summarize the split usage in~\cref{tab:12}. For each benchmark, the appearance encoder is pre-trained only on the training split, and the segmentation model is also trained on the corresponding training split. Evaluation is conducted on the validation split of each benchmark. We do not use cross-dataset pre-training for appearance distillation.

\begin{table}[h]
\centering
\scriptsize
\caption{\textbf{Preparation stage cost.} 
The preparation stage is performed offline. Scene-level steps are computed independently per scene, while teacher training is performed once per dataset.}
\label{tab:13}
\begin{tabular}{lcc}
\toprule
\textbf{Stage} & \textbf{Number of scenes} & \textbf{Runtime} \\
\midrule
COLMAP & 10 & 2.81 min / scene \\
GS Optimization~\cite{scenesplat} & 1613 & 22.00 min / scene \\
G2P Augmentation & 1201 & 11.01 s / scene \\
Boundary Extraction & 1201 & 0.09 s / scene \\
Teacher Training$^{\mathrm{a}}$ & 1201 & 27.9 h / dataset \\
\bottomrule
\end{tabular}
\begin{flushleft}
\scriptsize $^{\mathrm{a}}$ One-time cost per dataset.
\end{flushleft}
\end{table}

\subsubsection{Preparation cost and inference decoupling.}
\cref{tab:13} reports the full cost of the offline preparation stage, including GS reconstruction, G2P preprocessing, and appearance-encoder training. Since our experiments use Gaussian primitives from SceneSplat-7K and ScanNet provides depth measurements, COLMAP is not required in the actual G2P pipeline; we report its average runtime on 10 scenes only as a practical reference for generic image-based reconstruction pipelines. GS optimization, G2P augmentation, and boundary extraction are computed independently per scene, while the appearance encoder is trained once per dataset and then reused. Importantly, Gaussians are required only during this offline preparation stage. During inference, G2P operates as a Gaussian-free point-only segmentation model, so these preparation costs do not introduce additional overhead to the segmentation backbone. This positions G2P as an offline mapping framework for dense indoor scenes rather than a real-time replacement for raw point cloud segmentation.

\section{Observations and Analysis}
\label{sec:analysis}

\begin{figure}[t]
\centering
\includegraphics[width=\columnwidth]{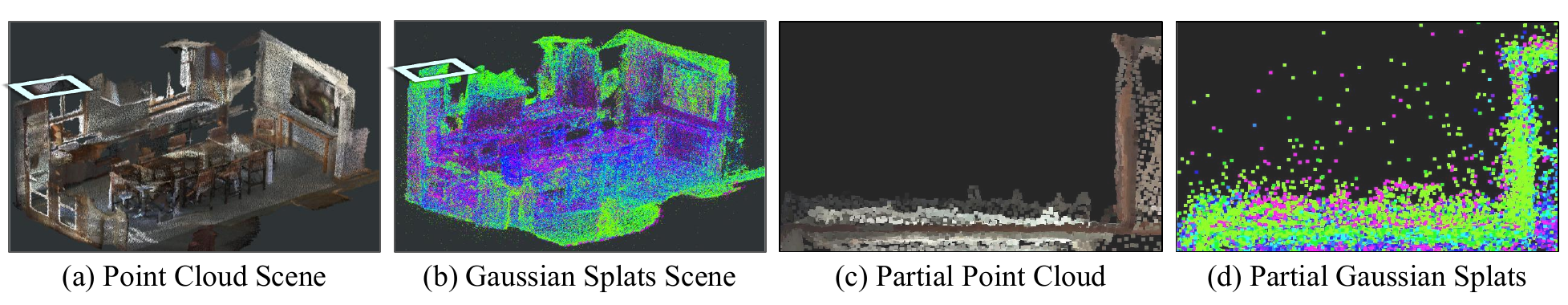}
\caption{\textbf{Comparison between point clouds and Gaussian splats.}
(a) Original point cloud scene.
(b) Reconstructed Gaussian splats of the same scene.
(c) A zoomed-in region from (a) in the white box.
(d) The corresponding region from (b), where boundary noise and geometric distortion appear.
Best viewed in color and zoom.}
\label{fig:6}
\end{figure}

This section covers: (i) point-based geometry preservation (\cref{fig:6}), (ii) comparison between Euclidean- and Mahalanobis-based G2P augmentation (\cref{fig:7}), (iii) scale and opacity analysis (\cref{fig:8}),  (iv) appearance feature analysis (\cref{fig:9}),  (v) Gaussian-segmentation quality correlation (\cref{fig:10}), and (vi) detailed G2P augmentation procedures (\cref{alg:1}). These analyses provide empirical support for the design choices of G2P and clarify how Gaussian-derived attributes contribute to segmentation.

\begin{algorithm}[t]
\caption{Gaussian-to-Point Feature Augmentation}
\label{alg:1}
\begin{algorithmic}[1]
\Require Point cloud $\mathcal{P} = \{p_i\}_{i=1}^N$, 3D Gaussians $\mathcal{G} = \{g_j\}_{j=1}^M$, maximum neighbors $k$, radius $r^g$
\Ensure Augmented point cloud $\mathcal{P}' = \{p_i'\}_{i=1}^N$
\State Initialize $\mathcal{P}' \gets \emptyset$ and pre-compute $\mathbf{\Sigma}_j^{-1}$ for all $g_j \in \mathcal{G}$
\For{$i = 1$ to $N$}
    \State Find candidates $\mathcal{G}_{\text{cand}} = \{ g_j \in \mathcal{G} \mid \|\mu_i^p - \mu_j^g\|_2 \le r^g \}$
    \State Select up to $k$ nearest Gaussians $\mathcal{G}_i$ and distances $D_i$ using Mahalanobis distance;
    \State \hspace{1.5em}if $\mathcal{G}_{\text{cand}} = \emptyset$, use the single Euclidean nearest Gaussian as fallback
    \State Compute normalized inverse-distance weights $w_{ij}$ from $D_i$ (Eq.~1)
    \State Aggregate features $S_i'$ and $\alpha_i'$ using weighted sum (Eq.~2)
    \State Augment point $p_i' \gets (\mu_i^p, c_i, n_i, S_i', \alpha_i')$
    \State Update $\mathcal{P}' \gets \mathcal{P}' \cup \{p_i'\}$ (Eq.~3)
\EndFor
\State \Return $\mathcal{P}'$
\end{algorithmic}
\end{algorithm}

\subsubsection{Geometric preservation in G2P.}
3D semantic segmentation requires preserving precise geometric structures, as clear object boundaries are essential for accurate prediction. However, 3D GS applies iterative adaptive density control, including prune, clone, and split operations, which alter the original point cloud structure. This leads to geometric inconsistencies such as blurred or noisy boundaries, as shown in~\cref{fig:6}(d). To address this issue, we propose G2P, which preserves the geometry of the original point cloud while transferring only the useful Gaussian attributes to points. In this way, the geometry remains point-based, and Gaussian cues are added as complementary features. The overall procedure is shown in~\cref{alg:1}.

\subsubsection{G2P augmentation details.}
\begin{figure*}[t!]
    \centering    
    \includegraphics[width=0.9\textwidth]{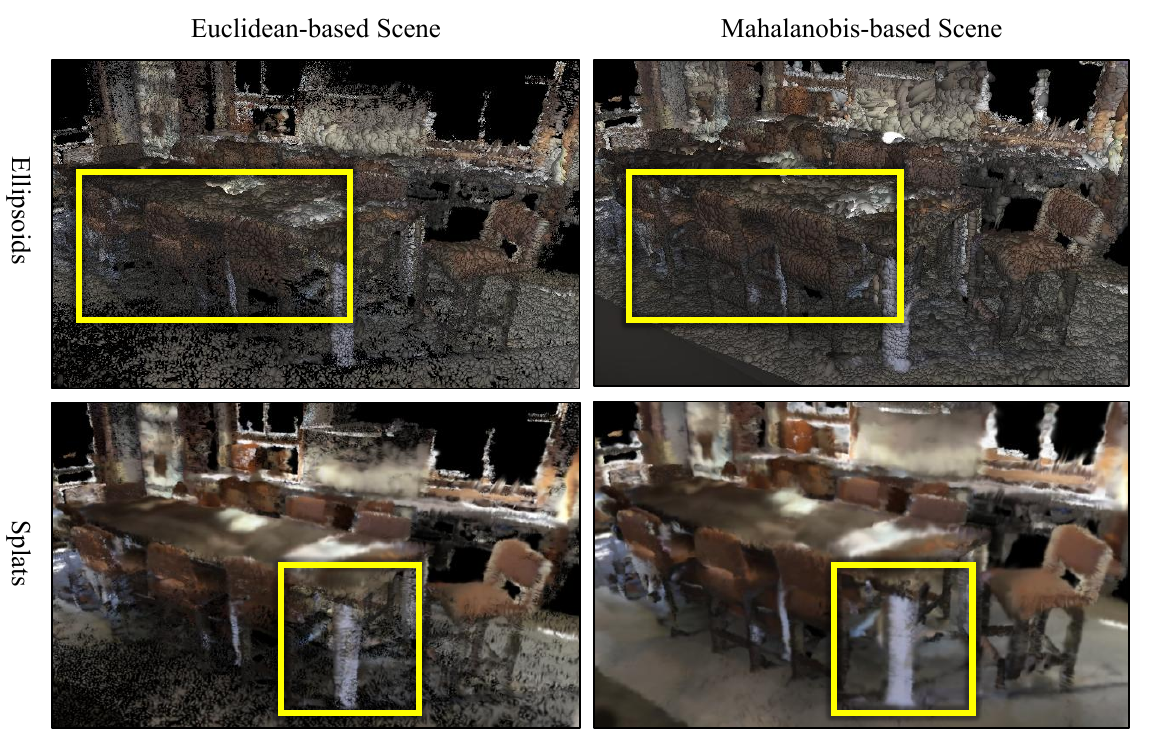}
    \caption{\textbf{Comparison of Euclidean and Mahalanobis-based Gaussian-to-Point augmentation.}
Left: Euclidean-based augmentation. Right: Mahalanobis-based augmentation. Top row: ellipsoid visualization. Bottom row: splats rendering.}
    \label{fig:7}
\end{figure*}

\cref{alg:1} summarizes the G2P augmentation process. For each point $p_i \in \mathcal{P}$, we identify candidate Gaussians within the search radius $r^g$ and rank them by Mahalanobis distance to select up to $k$ neighbors from $\mathcal{G}$. Their attributes are then aggregated to construct the augmented point $p_i'$. If no candidate Gaussian is found within $r^g$, we use the single Euclidean nearest Gaussian as a fallback. As shown in Tab.~10 of the main paper and \cref{fig:7}, Mahalanobis-based assignment yields higher mIoU than Euclidean matching and produces cleaner augmented structures (see yellow box).

\subsubsection{Empirical analysis of Gaussian attributes.} 
\begin{figure*}[t!]
    \centering    
    \includegraphics[width=0.9\textwidth]{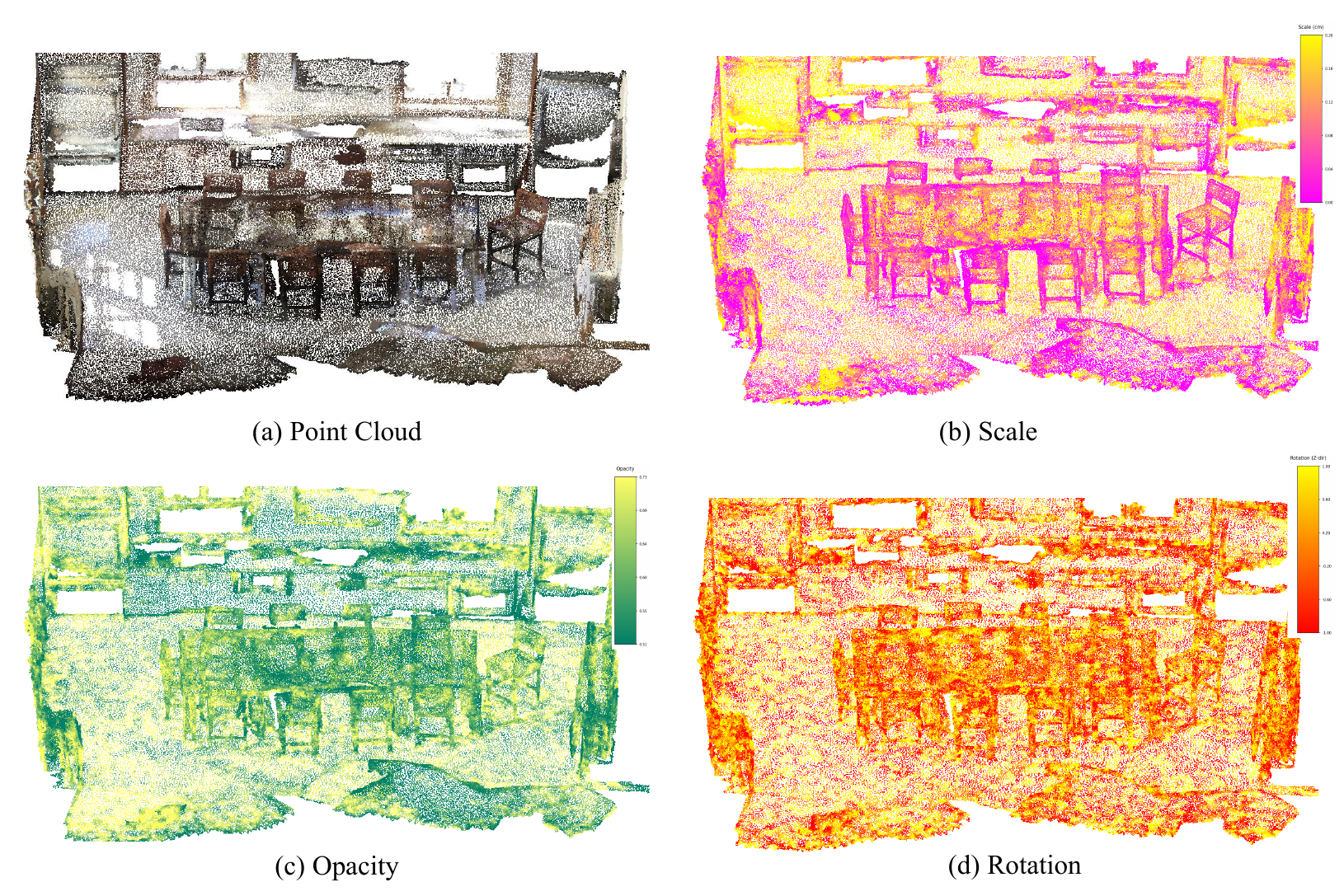}
    \caption{\textbf{Visualization of Gaussian attribute distributions.} 
(a) Point cloud. 
(b) \textbf{Scale:} The distribution clearly correlates with local geometry; small scales (pink) concentrate along object boundaries and fine-grained structures (e.g., chair legs), whereas large scales (yellow) dominate planar regions (e.g., floors). 
(c) \textbf{Opacity:} High opacity values indicate confident surface density. 
(d) \textbf{Rotation:} In contrast, rotation parameters exhibit a stochastic, unstructured pattern with limited correlation.}
    \label{fig:8}
\end{figure*}

In 3D GS~\cite{gaussiansplatting}, which has been recently adopted for large-scale scene understanding~\cite{scenesplat} and object representation learning~\cite{shapesplat}, a scene is represented by a set of anisotropic volumetric primitives. Each Gaussian is characterized by learnable attributes: geometric properties defined by the center position $\mu$ and covariance $\Sigma$, and visual properties encoded by opacity $\alpha$ and spherical harmonics (SH). Here, we provide empirical insights into how these attributes correlate with geometry and visibility-confidence signals. Interestingly, \cref{fig:8} reveals that \textbf{scale} and \textbf{opacity} provide complementary geometric and visibility-confidence cues for segmentation, while rotation exhibits noisy patterns with limited semantic correlation. 

We exclude SH coefficients and rotation parameters to prioritize robust structural learning over high-frequency noise. SH coefficients, being high-dimensional view-dependent, may interfere with the structural signals from scale and opacity; ShapeSplat~\cite{shapesplat} also shows that adding SH coefficients degrades segmentation performance compared to using only geometric attributes. Rotation (\cref{fig:8}(d)) exhibits stochastic distributions that do not correlate with object boundaries, unlike the structured patterns in scale and opacity. By restricting input to essential attributes (position, scale, and opacity), G2P encourages robust structural understanding rather than reliance on high-dimensional radiance-field descriptors.

\subsubsection{Scale as geometric guidance.} 
As visualized in~\cref{fig:8}(b), scale values correlate strongly with local geometry: smaller scales concentrate along object boundaries and thin structures, whereas larger scales dominate broad planar regions. Such distribution naturally emerges during 3D GS optimization, where small-scale Gaussians preserve sharp geometric discontinuities for rendering fidelity. This makes scale particularly useful for identifying boundaries in geometrically ambiguous regions where point-only cues are weak. Consequently, scale serves as an effective geometric cue for boundary detection, particularly for fine-grained objects adjacent to walls or floors.

\begin{figure*}[t!]
    \centering    
    \includegraphics[width=\textwidth]{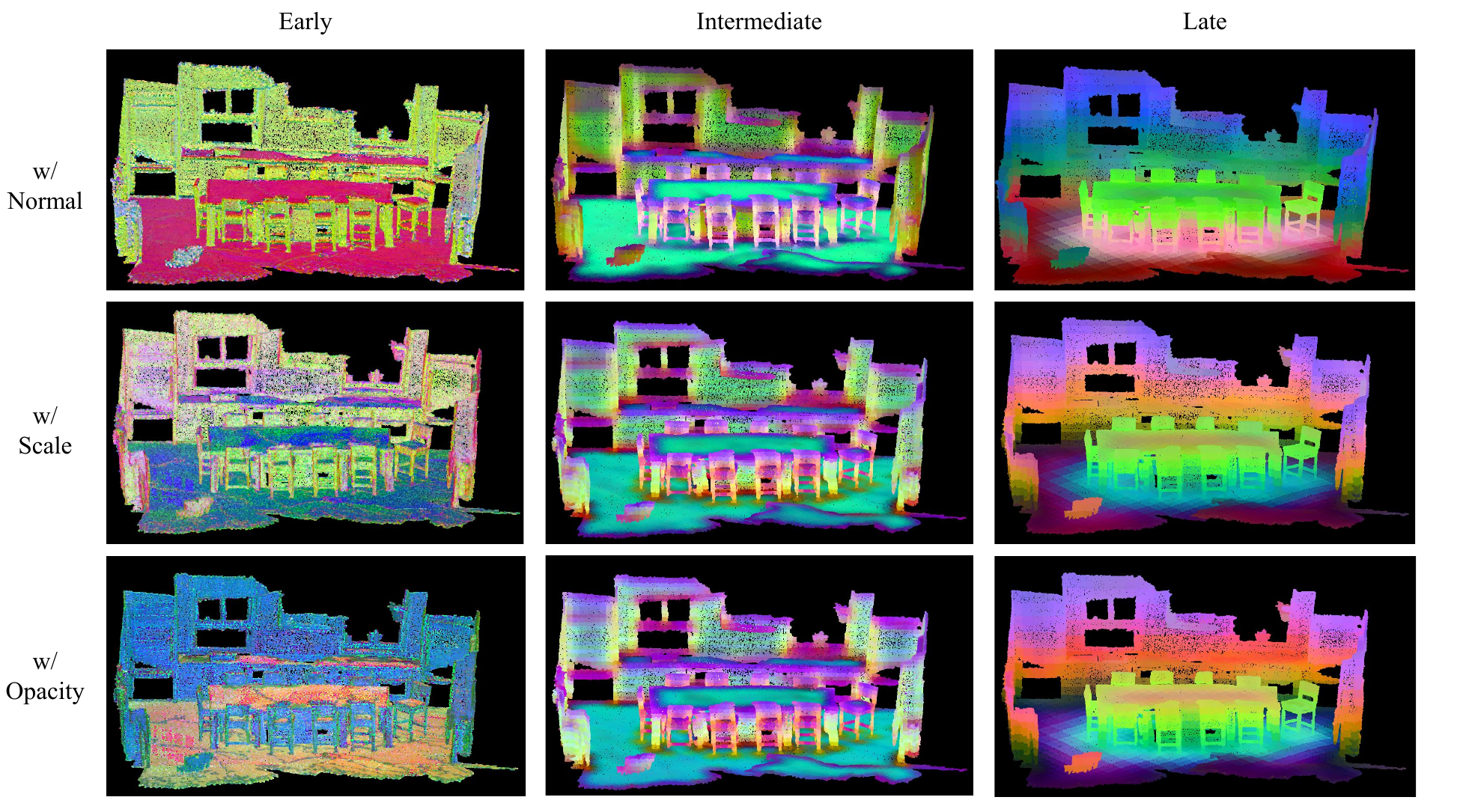}    
    \caption{\textbf{PCA visualization of appearance encoder features.}
Early, intermediate, and late features are shown for appearance encoders trained with different inputs: $(\mu^{p}, c, n)$, $(\mu^{p}, c, S')$, and $(\mu^{p}, c, \alpha')$.}
    \label{fig:9}
\end{figure*}

\subsubsection{Opacity as structural confidence.}
In contrast, opacity (\cref{fig:8}(c)) provides a visibility-confidence signal rather than a direct color or radiance descriptor. Background regions, such as walls and floors, exhibit uniformly high opacity (yellow) due to consistent multi-view visibility, while foreground objects display more varied distributions (green) reflecting geometric complexity and self-occlusion patterns. This view-invariant property arises from multi-view consistency optimization, making opacity a useful auxiliary cue. Prior work~\cite{shapesplat} demonstrates that opacity achieves low reconstruction error, confirming its discriminative power.

\subsubsection{Feature analysis of the appearance encoder.}
The backbone learns auxiliary visual cues via distillation from a GS-trained teacher. As shown in~\cref{fig:9}, teachers trained with GS-derived attributes organize features more distinctly than the normal-based variant. In the early stage, scale- and opacity-based encoders already exhibit more complex feature distributions, indicating that GS attributes provide cues beyond pure geometry. In the middle stage, these features become more spatially structured than those of the normal-based encoder. In the late stage, both GS-attribute variants show representations that are closer to semantic grouping, with the opacity-based encoder producing the clearest separation between regions, \eg, floor and chair. These structured features are distilled to the Gaussian-free student network, helping the backbone capture visual distinctions that are difficult to infer from geometry alone.

\begin{figure}[t]
\centering

\begin{minipage}[t]{0.48\columnwidth}
\centering
\includegraphics[width=\linewidth]{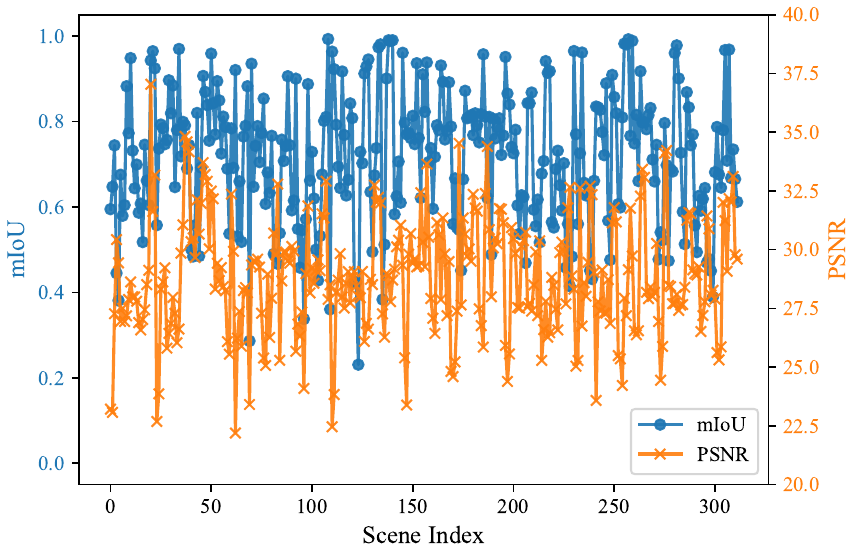}
\caption{\textbf{Scene-wise correlation between mIoU and PSNR.}
Scene-wise comparison between segmentation performance (mIoU) and Gaussian reconstruction quality (PSNR) on the ScanNet v2 validation set (312 scenes). 
Blue points indicate the mIoU of each scene, while orange points denote the average PSNR of its corresponding Gaussian reconstruction from SceneSplat-7K~\cite{scenesplat}.}
\label{fig:10}
\end{minipage}
\hfill
\begin{minipage}[t]{0.48\columnwidth}
\centering
\includegraphics[width=\linewidth]{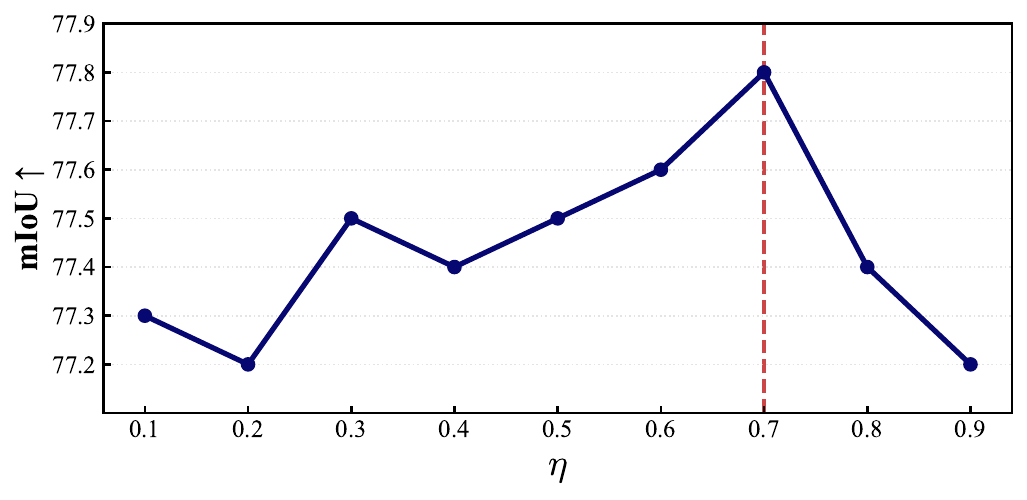}
\caption{\textbf{Ablation on pruning ratio $\eta$.}
The pruning ratio $\eta$ controls the scale-based component of the final boundary pseudo-labels 
$\mathcal{B} = \mathcal{B}_{\text{scale}} \cup \mathcal{B}_{\text{sem}}$.
Boundary guidance is evaluated without appearance distillation, with boundary loss weight $\lambda_b = 1.0$.
Results are reported in mIoU on the ScanNet v2 validation set.}
\label{fig:11}
\end{minipage}

\end{figure}

\subsubsection{Gaussian–segmentation quality correlation.}
As shown in~\cref{fig:10}, we analyze the relationship between scene-wise segmentation performance (\ie., mIoU) and Gaussian reconstruction quality (\ie., PSNR) across all 312 validation scenes in ScanNet v2~\cite{scannet}. Overall, the two metrics do not exhibit a strong correlation: scenes with low PSNR can still achieve high mIoU, and high PSNR does not necessarily guarantee better segmentation accuracy. However, we observe that extremely low-PSNR scenes (below 25 dB) tend to show degraded mIoU in some cases, which aligns with the PSNR-based filtering procedure noted in SceneSplat~\cite{scenesplat}. These observations suggest that G2P does not heavily rely on the photometric fidelity of Gaussians; instead, it primarily leverages the underlying point cloud geometry, with Gaussian attributes serving as complementary guidance.

\section{Ablation Studies}
\label{sec:ablation}

\begin{figure*}[t]
    \centering
    \includegraphics[width=\textwidth]{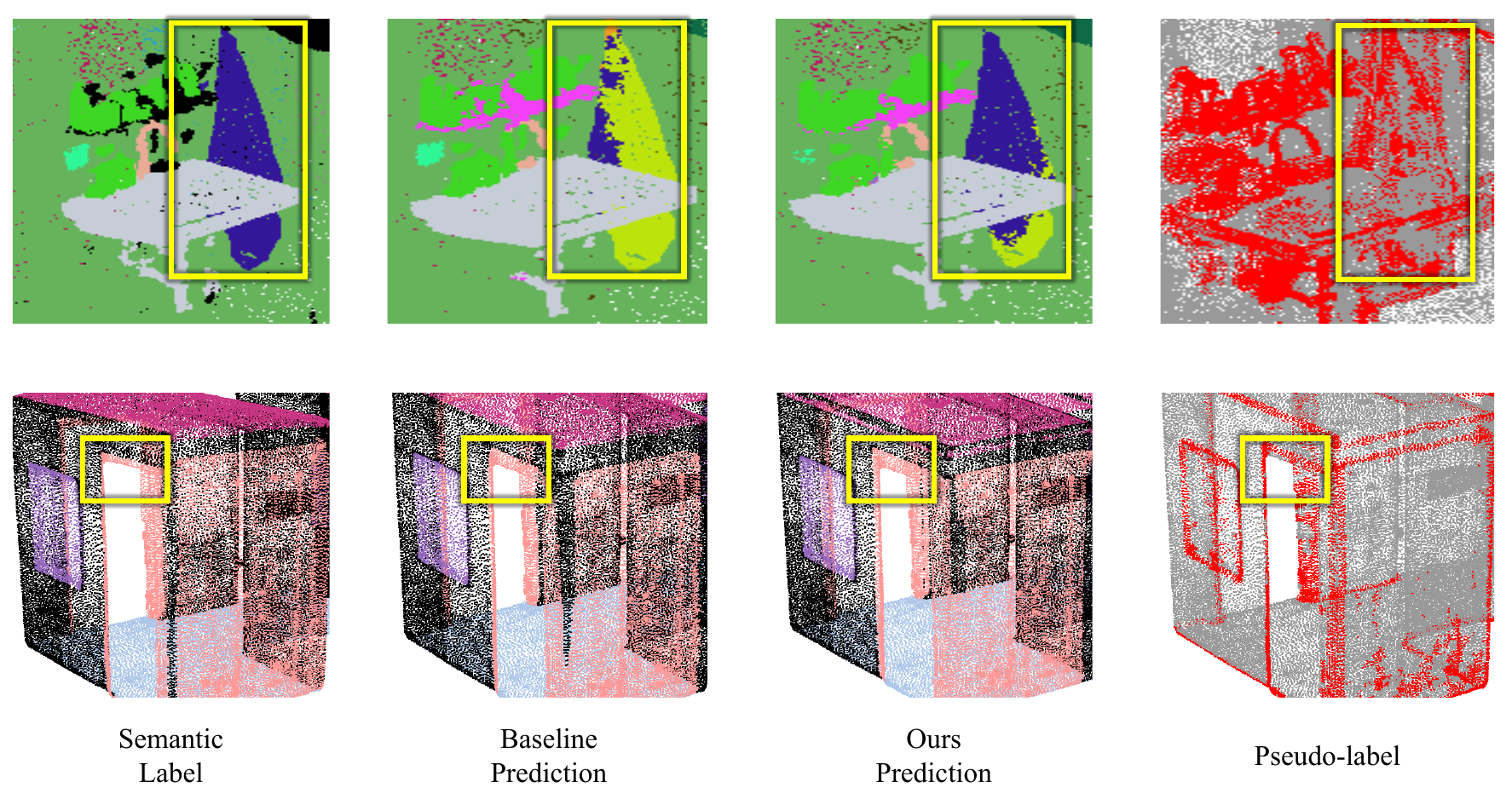}
\caption{\textbf{Qualitative results on ScanNet++ and Matterport3D.}
Top row: ScanNet++ scene; bottom row: Matterport3D scene. From left to right: ground-truth semantic labels, PT v3 baseline predictions, G2P predictions, and scale-based boundary pseudo-labels. Yellow boxes highlight challenging objects where G2P produces clearer boundaries and more accurate segmentation compared to the baseline.}
    \label{fig:12}
\end{figure*}

This section validates the main design choices of G2P through supplementary ablations and analyses, including: (i) the effectiveness of GS-derived attributes (\cref{tab:14}), (ii) the sensitivity of the scale-based boundary pruning ratio $\eta$ (\cref{fig:11}), (iii) the sensitivity of the auxiliary loss weights (\cref{tab:15}), (iv) the trade-off induced by appearance distillation across geometry-dependent and appearance-challenging classes (\cref{tab:16}), and (v) the effectiveness of object-focused scale boundary supervision (\cref{tab:wo_obj_boundary}). Additional qualitative comparisons on ScanNet++ and Matterport3D are also provided in \cref{fig:12}. All experiments use the ScanNet v2 validation set unless noted otherwise.

\begin{table}[t]
\centering
\scriptsize
\setlength{\tabcolsep}{3pt}
\renewcommand{\arraystretch}{0.95}
\caption{\textbf{Effectiveness of GS-derived attributes under direct input and distillation settings.} \textit{Backbone} denotes directly training the segmentation backbone with the corresponding input features, whereas \textit{Distillation} denotes pre-training an appearance encoder with Sonata using the same inputs and distilling it to the backbone without the B--S block. $\mathrm{SH}'$ denotes aggregated spherical harmonics (degree=1) from G2P augmentation, and $^{\dagger}$ indicates reproduction by us.}
\label{tab:14}
\begin{tabular}{llccc}
\toprule
\textbf{Type} & \textbf{Input features} & \textbf{mIoU$\uparrow$} & \textbf{mAcc$\uparrow$} & \textbf{OA$\uparrow$} \\
\midrule
\multirow{8}{*}{Backbone}
& $(\mu^{p}, c, n)$ (PT v3$^{\dagger}$) & 77.0 & 84.3 & 92.1 \\
& $(\mu^{p}, c, S')$ & \textbf{77.4} & 85.1 & \textbf{92.2} \\
& $(\mu^{p}, c, \alpha')$ & 77.2 & 84.8 & 91.8 \\
& $(\mu^{p}, c, \mathrm{SH}')$ & 77.1 & 84.7 & 92.0 \\
& $(\mu^{p}, c, S', \mathrm{SH}')$ & \textbf{77.4} & \textbf{85.6} & 92.0 \\
& $(\mu^{p}, c, S', \alpha')$ & 77.3 & 84.5 & \textbf{92.2} \\
& $(\mu^{p}, c, \mathrm{SH}', \alpha')$ & 77.0 & 84.5 & 91.9 \\
& $(\mu^{p}, c, S', \mathrm{SH}', \alpha')$ & 77.2 & 85.1 & 92.1 \\
\midrule
\multirow{5}{*}{Distillation}
& $(\mu^{p}, c, n)$ & 76.5 & 83.7 & 91.6 \\
& $(\mu^{p}, c, S')$ & 77.3 & 84.2 & 91.8 \\
& $(\mu^{p}, c, \mathrm{SH}')$ & 77.0 & 83.9 & 91.6 \\
& $(\mu^{p}, \alpha')$ & 77.4 & 85.0 & 92.0 \\
& \cellcolor{gray!12}{$(\mu^{p}, c, \alpha')$} & \cellcolor{gray!12}{77.8} & \cellcolor{gray!12}{\textbf{85.4}} & \cellcolor{gray!12}\textbf{{92.1}} \\
& $(\mu^{p}, c, S', \alpha')$ & \textbf{78.0} & 85.1 & \textbf{92.1} \\
\bottomrule
\end{tabular}
\end{table}

\begin{table}[t]
\centering
\caption{\textbf{Ablation on loss weights.} We vary the appearance distillation weight $\lambda_d$ and the boundary guidance weight $\lambda_b$, while keeping the other fixed at its best-performing value ($\lambda_d = 0.4$, $\lambda_b = 0.9$).}
\label{tab:15}
\scriptsize
\setlength{\tabcolsep}{2pt}
\renewcommand{\arraystretch}{1.0}
\begin{tabular}{@{}lccc|lccc@{}}
\toprule
\multicolumn{4}{c|}{\textbf{Appearance distillation}} & \multicolumn{4}{c}{\textbf{Boundary guidance}} \\
\cmidrule(lr){1-4} \cmidrule(lr){5-8}
\textbf{$\lambda_d$} & \textbf{mIoU$\uparrow$} & \textbf{mAcc$\uparrow$} & \textbf{OA$\uparrow$} & \textbf{$\lambda_b$} & \textbf{mIoU$\uparrow$} & \textbf{mAcc$\uparrow$} & \textbf{OA$\uparrow$} \\
\midrule
0.1 & 76.8 & 84.2 & 91.9 & 0.1 & 76.8 & 84.3 & 91.9 \\
0.2 & 77.8 & \textbf{85.3} & 92.0 & 0.5 & 77.0 & 84.4 & 91.8 \\
0.3 & 77.4 & 84.8 & 91.9 & 0.6 & 77.3 & 84.5 & 92.0 \\
\cellcolor{gray!12}\textbf{0.4} & \cellcolor{gray!12}\textbf{78.4} & \cellcolor{gray!12}85.2 & \cellcolor{gray!12}\textbf{92.4} & 0.7 & 77.1 & 84.4 & 91.8 \\
0.5 & 77.4 & 84.2 & 91.9 & 0.8 & 77.9 & \textbf{85.8} & 92.2 \\
0.6 & 77.3 & 85.0 & 91.9 & \cellcolor{gray!12}\textbf{0.9} & \cellcolor{gray!12}\textbf{78.4} & \cellcolor{gray!12}85.2 & \cellcolor{gray!12}\textbf{92.4} \\
0.7 & 77.0 & 84.1 & 92.0 & 1.0 & 77.9 & 85.3 & 92.1 \\
\bottomrule
\end{tabular}
\end{table}

\subsubsection{Effectiveness of GS-derived attributes.}
\cref{tab:14} shows that GS-derived attributes consistently improve over the PT v3 baseline, indicating that Gaussian cues provide useful information beyond standard point features for 3D scene perception. When used as direct backbone inputs, scale yields the most consistent gains, whereas spherical harmonics provide relatively limited improvements. More importantly, distillation is generally more effective than direct input, with opacity showing the strongest single-attribute performance and the combination of scale and opacity achieving the best overall result. In contrast, distilling from a teacher trained only on conventional point attributes $(\mu^{p}, c, n)$ leads to noticeably lower performance, suggesting that the gain does not come from distillation alone. Rather, the improvement comes from learning GS-derived visual cues and transferring them through distillation.

\subsubsection{Sensitivity of Scale-based boundary pruning ratio $\eta$.}
As shown in~\cref{fig:11}, performance varies smoothly across different values of $\eta$, with the best mIoU obtained at $\eta = 0.7$. This trend suggests that moderate pruning yields cleaner boundary candidates by suppressing large-scale planar regions while preserving informative object contours. We therefore use $\eta = 0.7$ in our main experiments.

\subsubsection{Loss weight sensitivity.}
As shown in~\cref{tab:15}, performance changes smoothly across different values of $\lambda_d$ and $\lambda_b$, with the best mIoU achieved at $\lambda_d = 0.4$ and $\lambda_b = 0.9$. The relatively smooth variation indicates that G2P is not overly sensitive to the exact weighting of the auxiliary losses, while still benefiting from balanced appearance distillation and boundary supervision. These settings are therefore adopted in our main experiments.

\subsubsection{Qualitative results on ScanNet++ and Matterport3D.}
Tab.~6 in the main paper shows that G2P improves segmentation over the PT v3 baseline on both ScanNet++ and Matterport3D. \cref{fig:12} qualitatively illustrates these gains. The PT v3 baseline often produces blurred or fragmented boundaries around thin structures and planar intersections, leading to misclassification between adjacent surfaces. In contrast, G2P generates more coherent predictions that better align with object boundaries. The scale-derived boundary pseudo-labels further show that small-scale Gaussians concentrate around object contours and thin structures, providing reliable cues for boundary-aware segmentation.

\begin{table}[t]
\centering
\scriptsize
\setlength{\tabcolsep}{1.5pt}
\caption{\textbf{Effect of the distillation weight $\lambda_d$ on geometry-dependent and appearance-challenging classes.}
Chair, Table, and Desk are geometry-dependent classes, whereas Refrigerator, Shwr.\ (Shower Curtain), and Window are appearance-challenging classes. Fine Avg and Chall. Avg denote the average IoU of the geometry-dependent and appearance-challenging groups, respectively.}
\begin{tabular}{c|ccc|c|ccc|c}
\toprule
 & \multicolumn{4}{c|}{\textbf{Geometry-dependent}} & \multicolumn{4}{c}{\textbf{Appearance-challenging}} \\
\cmidrule(lr){2-5} \cmidrule(lr){6-9}
$\lambda_d$ & Chair & Table & Desk & Fine Avg & Refrigerator & Shwr. & Window & Chall. Avg \\
\midrule
0.2 & 93.34 & 79.46 & 74.43 & \textbf{82.41} & 69.52 & 70.65 & 72.41 & 70.86  \\
0.4 & 92.67 & 79.07 & 74.52 & 82.09 & 71.35 & 74.04 & 72.24 & 72.54  \\
0.8 & 91.94 & 78.43 & 71.37 & 80.58 & 70.15 & 73.52 & 75.36 & \textbf{73.01}  \\
\bottomrule
\end{tabular}
\label{tab:16}
\end{table}

\begin{table}[t]
\centering
\scriptsize
\setlength{\tabcolsep}{3pt}
\renewcommand{\arraystretch}{0.92}
\small
\caption{\textbf{Boundary supervision ablation with sources for $\mathcal{B}_{\mathrm{scale}}$.} $\mathcal{B}_{\mathrm{sem}}$ denotes semantic boundary supervision with radius $r^s$, and $\mathcal{B}_{\mathrm{scale}}$ denotes scale-based boundary supervision with trimming ratio $\eta$. For $\mathcal{B}_{\mathrm{scale}}$, we compare extracting boundary points from $\mathcal{P}'$ and from the object-focused subset $\mathcal{P}'_{\mathrm{obj}}$, following Eq.~(5). All results are obtained without appearance distillation. $^{\dagger}$ indicates reproduction by us.}
\label{tab:wo_obj_boundary}
\begin{tabular}{lcccc}
\toprule
\textbf{Boundary supervision} & \textbf{Sem. label} & \textbf{mIoU$\uparrow$} & \textbf{mAcc$\uparrow$} & \textbf{OA$\uparrow$} \\
\midrule
No Supervision (PT v3$^{\dagger}$) &  & 77.0 & 84.3 & 92.1 \\
\midrule
$\mathcal{B}_{\mathrm{sem}} \ (r=0.04)$ & \checkmark & 76.9 & 84.3 & 91.8 \\
\rowcolor{gray!12}
$\mathcal{B}_{\mathrm{sem}} \ (r=0.04) + \mathcal{B}_{\mathrm{scale}} \ (p_i' \in \mathcal{P}'_{\mathrm{obj}},\ \eta=0.7)$ & \checkmark & 77.8 & 85.4 & 92.1 \\
\midrule
$\mathcal{B}_{\mathrm{scale}} \ (p_i' \in \mathcal{P}',\ \eta=0.3)$ &  & 76.3 & 84.0 & 91.7 \\
$\mathcal{B}_{\mathrm{scale}} \ (p_i' \in \mathcal{P}',\ \eta=0.5)$ &  & 76.8 & 84.5 & 91.8 \\
$\mathcal{B}_{\mathrm{scale}} \ (p_i' \in \mathcal{P}',\ \eta=0.7)$ &  & 77.0 & 84.1 & 91.8 \\
$\mathcal{B}_{\mathrm{scale}} \ (p_i' \in \mathcal{P}',\ \eta=0.9)$ & & \underline{77.5} & \underline{84.8} & \underline{92.1} \\
\bottomrule
\end{tabular}
\end{table}

\subsubsection{Trade-off between geometry-dependent and appearance-challenging classes.}
To better understand the effect of appearance distillation, we analyze the impact of the distillation weight $\lambda_d$ on two class groups defined in~\cref{tab:16}. As shown in~\cref{tab:16}, increasing $\lambda_d$ improves performance on appearance-challenging classes, while slightly reducing accuracy on geometry-dependent classes. This trend indicates that stronger appearance distillation helps resolve ambiguities between objects with similar geometry but distinct appearance, at the cost of slightly weaker fine-grained geometric discrimination. At the same time, the degradation on geometry-dependent classes remains relatively small, suggesting that appearance cues mainly benefit categories where geometric evidence alone is insufficient. In practice, we select $\lambda_d = 0.4$ as a balanced setting that preserves strong overall performance while improving robustness to geometric ambiguity.

\subsubsection{Ablations on object-focused scale boundary supervision.}
\cref{tab:wo_obj_boundary} compares different sources of $\mathcal{B}_{\mathrm{scale}}$. In our method, extracting $\mathcal{B}_{\mathrm{scale}}$ from the object-focused subset $\mathcal{P}'_{\mathrm{obj}}$ yields the strongest results, indicating that concentrating boundary cues on object regions is more effective than using all points. This is likely because large planar background regions dominate the scale distribution and weaken object-level boundary signals. 

The scale-only results further show that GS-derived scale cues remain effective even when $\mathcal{B}_{\mathrm{scale}}$ is extracted directly from $\mathcal{P}'$, and performance improves as $\eta$ increases. This suggests that removing more large-scale points leaves cleaner boundary candidates. Notably, even without semantic filtering, the scale-only setting at $\eta=0.9$ remains competitive, indicating that Gaussian scale itself provides a meaningful boundary prior. Overall, while object-focused filtering gives the strongest supervision, GS-derived scale cues remain effective even without semantic labels.

\section{Additional Visualizations}
\label{sec:add_Vis}

\begin{figure*}[t]
    \centering
    \includegraphics[width=\textwidth]{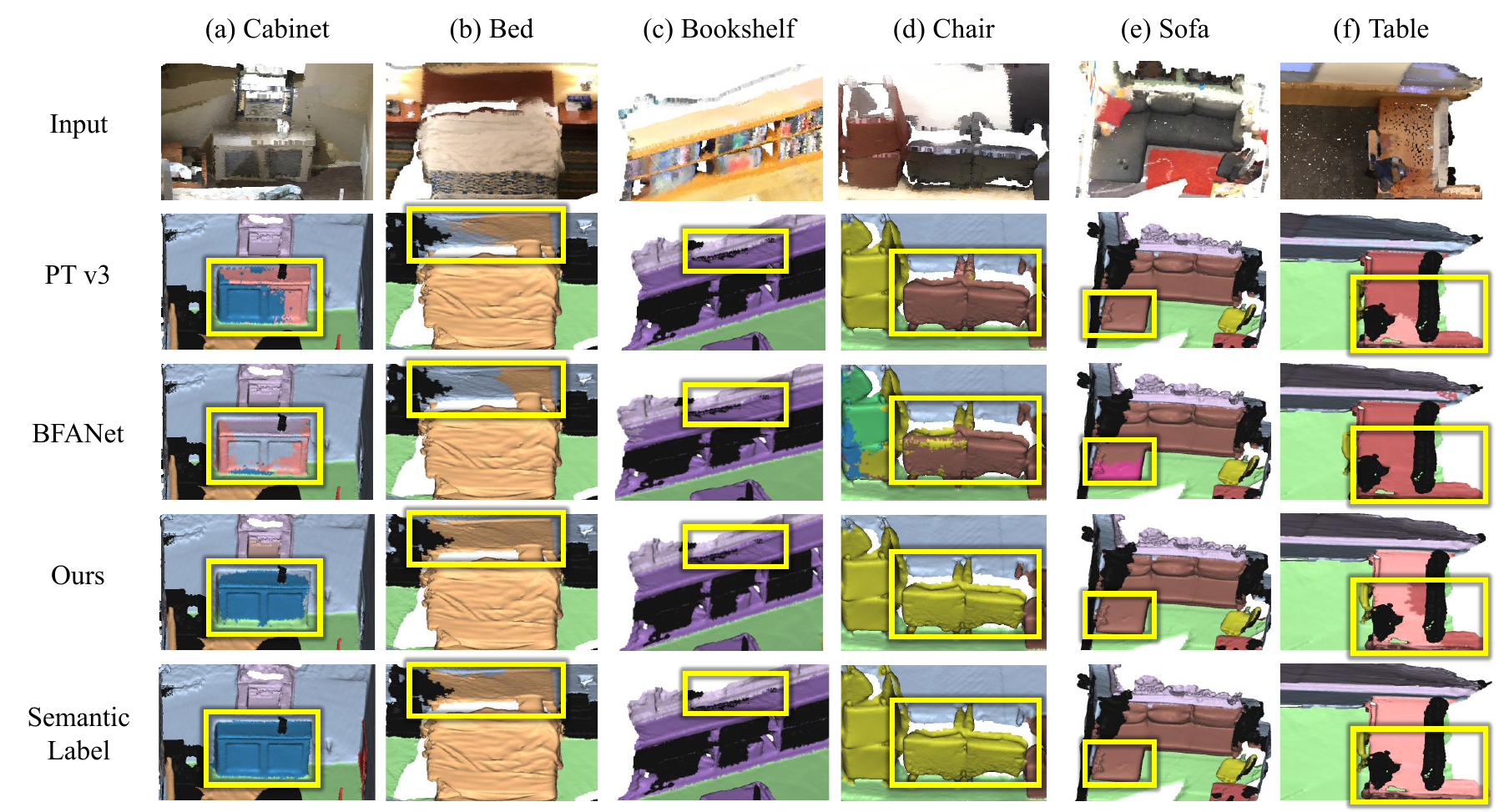}
    \caption{\textbf{Qualitative comparison on distinguishable classes in ScanNet v2.}
    G2P accurately separates objects sharing similar color or texture with their surroundings, 
    highlighting improved distinction over baselines in ambiguous regions.}
    \label{fig:13}
\end{figure*}

\subsubsection{Additional visualizations on ScanNet v2.} \cref{fig:13} shows results on six representative classes. For geometrically distinct objects (columns (a), (b), (c): cabinet, bed, bookshelf), G2P accurately captures complete object regions, while PT v3~\cite{PointTFv3} and BFANet~\cite{bfanet} either leak into adjacent structures or leave large portions unlabeled. For objects with geometric ambiguity (columns (d), (e), (f): chair, sofa, table), baseline methods merge surfaces with similar-looking backgrounds despite clear geometric boundaries.

\subsubsection{Boundary pseudo-label visualization.}
\begin{figure*}[!t]
    \centering
    \includegraphics[width=\textwidth]{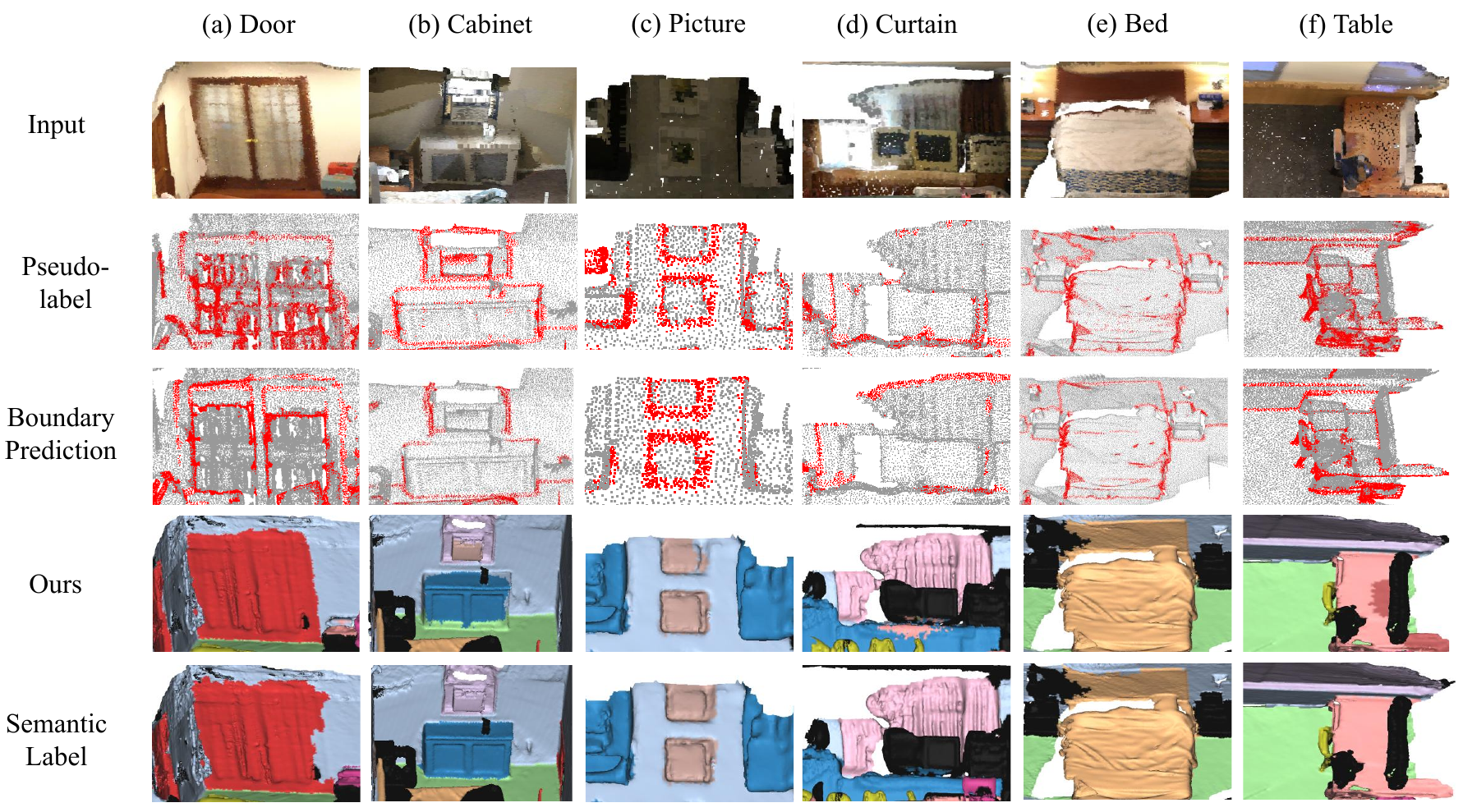}
    \caption{\textbf{Boundary pseudo-labels and predictions.}
    Columns correspond to different object categories, while rows show the input, scale-based boundary pseudo-labels, boundary predictions from G2P, final segmentation results, and ground-truth semantic labels.}
    \label{fig:14}
\end{figure*}

\cref{fig:14} shows that scale-based pseudo-labels derived from Gaussian scale attributes closely follow object contours across diverse categories. Geometrically challenging objects (columns (a), (c), (d): door, picture, curtain) exhibit denser boundary responses than more distinguishable objects (columns (b), (e), (f): cabinet, bed, table), confirming that small-scale Gaussians concentrate at high-frequency edges, as observed in \cref{fig:8}(b). 

\begin{figure*}[t]
    \centering
    \includegraphics[width=\textwidth]{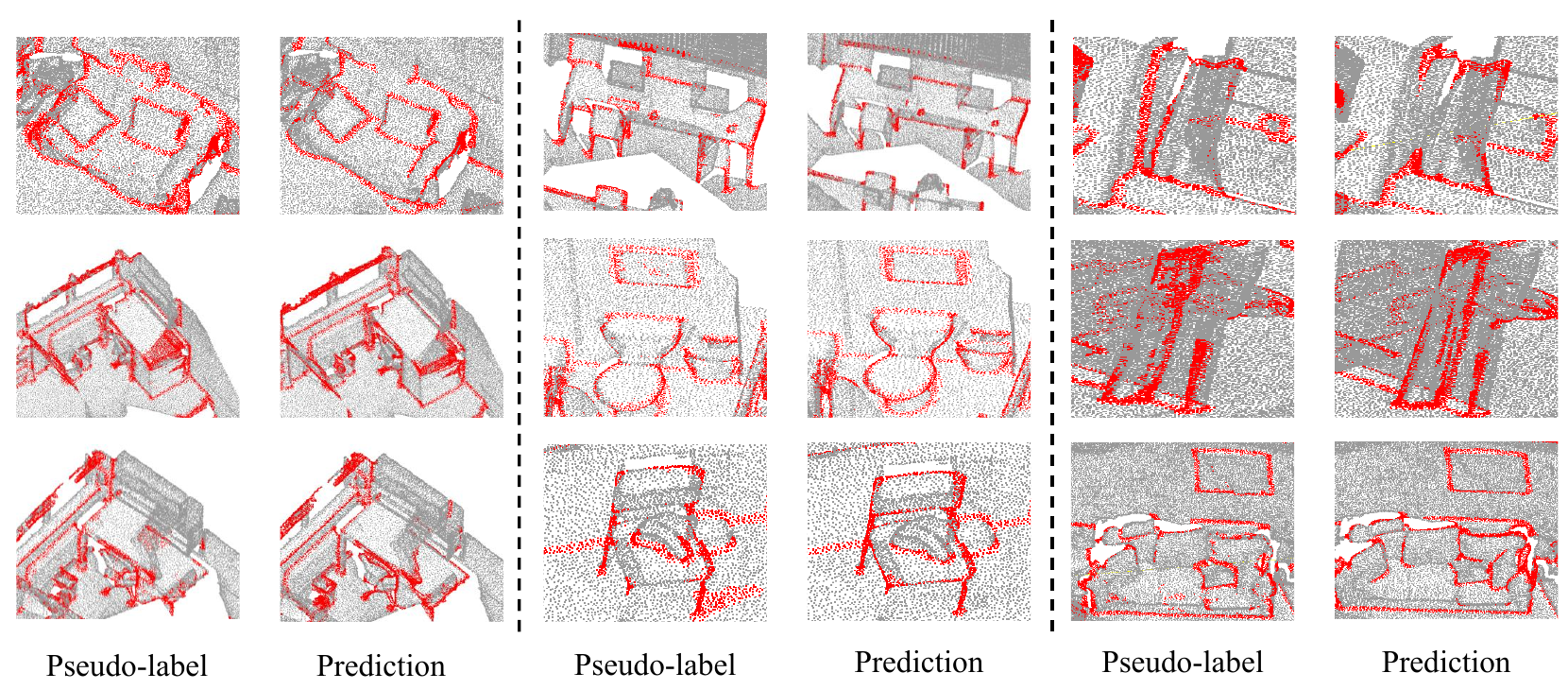}
\caption{\textbf{Additional boundary pseudo-label examples.}
    Left column in each pair: scale-based boundary pseudo-labels. 
    Right column: corresponding boundary predictions from G2P's boundary head, showing dense and precise localization along object edges across diverse scenes.}
    \label{fig:15}
\end{figure*}

Additional examples in~\cref{fig:15} further demonstrate that these pseudo-labels remain well aligned with the boundary predictions across indoor scenes. In particular, textured regions can induce small-scale Gaussians, as shown in~\cref{fig:14}(a) and~\cref{fig:15} (row~2, col.~2; \textit{picture above the toilet}), but the resulting predictions successfully suppress such artifacts through joint semantic-boundary supervision. Overall, these visualizations support that GS-derived scale cues provide reliable boundary priors even in challenging and cluttered scenes.

\section{Limitations}
\label{sec:limitation}
Outdoor scenes are significantly more challenging for current 3D GS-based representations. Since G2P relies on Gaussian scale statistics for boundary extraction, extending it to outdoor environments may require locally normalized scale statistics to reduce sensitivity to large-scale variations and outliers. Moreover, as discussed in \cref{tab:13}, the G2P pipeline operates as an offline preprocessing stage for dense indoor scenes. Recent lightweight or sparse-view 3D GS methods (\eg, DNGaussian~\cite{dngaussian} and Speedy-Splat~\cite{speedy-splat}) may help reduce this cost and enable more efficient extensions in future work.

\clearpage

\bibliographystyle{splncs04}
\bibliography{main}

\end{document}